\documentclass{article}

\usepackage[english]{babel}
\usepackage[round]{natbib}
\renewcommand{\bibname}{References}

\usepackage{wrapfig}
\usepackage{graphicx} % Required for inserting images
\usepackage[utf8]{inputenc} % allow utf-8 input
\usepackage[T1]{fontenc}    % use 8-bit T1 fonts
\usepackage{url}            % simple URL typesetting
\usepackage{booktabs}       % professional-quality tables
\usepackage{amsfonts}       % blackboard math symbols
\usepackage{nicefrac}       % compact symbols for 1/2, etc.
\usepackage{float} 
\usepackage{microtype}      % microtypography
\usepackage{xcolor}         % colors
\usepackage{amsmath} 
\usepackage{comment}
\usepackage{subcaption}  % subfigure support
\usepackage{multirow}
\usepackage{bbm}
\usepackage{tikz}              % core graphics package
\usetikzlibrary{bayesnet}  
\usepackage{amssymb}
\usepackage{multirow}
\usepackage{array,tabularx} % array gives \newcolumntype
\newcolumntype{C}[1]{>{\centering\arraybackslash}p{#1}}
\usepackage{authblk}
\usepackage[letterpaper,top=2cm,bottom=2cm,left=2.5cm,right=2.5cm,marginparwidth=1.75cm]{geometry}

\usepackage[colorlinks=true, allcolors=blue]{hyperref}

\title{Large-Scale Bayesian Causal Discovery with Interventional Data}

\author[1]{Seong Woo Han}
\author[2]{Daniel Duy Vo}
\author[1,3,4]{Brielin C. Brown\thanks{Correspondence: brielin@upenn.edu}}

\affil[1]{Department of Computer and Information Science, University of Pennsylvania}
\affil[2]{Genomics and Computational Biology Graduate Group, University of Pennsylvania}
\affil[3]{Division of Informatics, University of Pennsylvania}
\affil[4]{Department of Genetics, University of Pennsylvania}

\begin{document}
\maketitle

\begin{abstract}
Inferring the causal relationships among a set of variables in the form of a directed acyclic graph (DAG) is
an important but notoriously challenging problem. Recently, advancements in high-throughput genomic perturbation screens have inspired development of methods that leverage interventional data to improve model identification. However, existing methods still suffer poor performance on large-scale tasks  and fail to quantify uncertainty. Here, we propose Interventional Bayesian Causal Discovery (IBCD), an empirical Bayesian framework for causal discovery with interventional data. Our approach models the likelihood of the matrix of total causal effects, which can be approximated by a matrix normal distribution, rather than the full data matrix. We place a spike-and-slab horseshoe prior on the edges and separately learn data-driven weights for scale-free and Erdős–Rényi structures from observational data, treating each edge as a latent variable to enable uncertainty-aware inference. Through extensive simulation, we show that IBCD achieves superior structure recovery compared to existing baselines. We apply IBCD to CRISPR perturbation (Perturb-seq) data on 521 genes, demonstrating that edge posterior inclusion probabilities enable identification of robust graph structures.
\end{abstract}

\section{Introduction}
Inferring the causal relationships among a set of measured variables is one of the fundamental challenges
of science~\citep{bunge_causality_2012}. % This is particularly true in genetics?
These relationships are typically represented in a directed acyclic graph (DAG), with nodes representing
variables and the presence of an edge from one variable to another indicating the former has a direct causal effect on the latter. Unfortunately the DAG is not identifiable from observational data alone~\citep{Verma1990}, and
identifying the equivalence class of DAGs consistent with the data is NP-hard, making it intractable for problems involving even a modest number of variables~\citep{Chickering1996}.

This issue has recently received substantial attention in genomics, where progress in understanding biological mechanisms
of disease is now thought to hinge on understanding gene regulatory networks containing thousands of variables~\citep{Boyle2017,Liu2019,Wray2018}.
Fortunately, advances in CRISPR screening techniques such as Perturb-seq~\citep{Dixit2016, Gasperini2019, Montalbano2017}
allow researchers to systematically perturb specific genes and measure the response, yielding 
\textit{interventional} data for hundreds to thousands of genes measured in upwards of one-million cells~\citep{, Replogle2022,feng_genome-scale_2024}. This has created an ideal setting for the development of methods for inferring
causal graphs at scale~\citep{Xue2023, Yang2018, brown2023large, nazaret2023stable, annadani2023bayesdag, hagele2023bacadi, weinstock_gene_2024}.
% Interventional data improve the identifiability of the DAG~\citep{Hauser2012},

While differentiable causal discovery methods~\citep{zheng2018dags, brouillard2020differentiable, nazaret2023stable, lopez2022large}
have made tremendous advancements and received substantial attention, they are limited to returning a single
optimal DAG without uncertainty quantification. In contrast, Bayesian methods~\citep{madigan1995bayesian,
geiger_parameter_2002, friedman_being_2003, agrawal_minimal_2018, hagele2023bacadi, annadani2023bayesdag, weinstock_gene_2024}
aim to characterize the posterior distribution over possible graphs. This is critical
% when data are high-dimensional and noisy,
as there may be multiple different near-optimal DAGs. Uncertainty is also
invaluable for scientific applications
where assessing the degree of confidence in an edge or graph structure is required in order to prioritize
follow-up investigations or experiments.

Due to the exponential scaling of the number of DAGs, exact posterior calculation is intractable for all
but the smallest problems~\citep{tian_computing_2009}. Most practical approaches have therefore focused on
using Markov Chain Monte Carlo (MCMC) to sample from the posterior~\citep{geiger_parameter_2002,friedman_being_2003,
agrawal_minimal_2018}, however these approaches do not accommodate intervention data. Recent work uses gradient
information to further speed up inference~\citep{annadani2023bayesdag,lorch2021dibs,cundy2021bcd, deleu2022bayesian, toth2022active, toth2024effective}, but these approaches still fail to scale to the volume of data we consider here.

We take an alternative approach. Rather than % model the distribution of the direct effect matrix given the data,
directly modeling the distribution of the DAG structure or direct edge weights given the data,
we use the data to calculate the total causal effect (TCE) of every feature on every other. This 
reduces the amount of data that needs to be considered when calculating the likelihood.
We then approximate the posterior distribution
of the direct effect matrix given the total effect matrix. While modeling direct effects from total
effects has been previously considered~\citep{hyttinen_learning_2012,brown2023large,weinstock_gene_2024},
we present the first fully Bayesian treatment of this model and the first causal discovery method
that scales to large problems while quantifying uncertainty. Moreover, while recent causal discovery approaches
often model interventions, they typically assume \textit{hard} or \textit{structural} interventions which alter the
DAG. We instead consider more
realistic \textit{soft} interventions, sometimes called \textit{shift} interventions~\citep{sani_identification_2020}, that perturb
targeted variables while leaving the structure in tact.
We summarize our contributions as follows:

\begin{itemize}
    \item We formulate the problem of inferring the DAG as finding the inverse of the TCE matrix in the linear-autoregressive model using soft interventions.
    
    \item We show that asymptotically normal estimators of the TCE entries lead to a sampling distribution that
    can be well-approximated by a matrix normal distribution. This substantially reduces the dimensionality of the sampling covariance and enables efficient likelihood-based inference.
    
    \item We propose a hybrid empirical Bayesian prior over the DAG by combining global sparsity patterns with edge specific weights optimized from covariance estimates in the observational data. We learn Erdős–Rényi (ER) and scale-free (SF) priors separately through structure-specific optimization, yielding a flexible, data-adaptive prior.
    
    \item We perform extensive empirical analyses while varying properties of the DAG, demonstrating superior performance to existing approaches combined with the ability to quantify uncertainty in edge inclusion.
    
    % \item We assess uncertainty on real data by cross validating perturbed expression profiles for 521 genes shared by the 
    % genome wide Perturb seq and Essential K562 screens~\citep{Replogle2022} and also by fitting once to each full screen, 
    % showing that posterior inclusion probabilities (PIPs) provide robust and practical measures of edge significance across 
    % folds and across datasets.
    \item We assess uncertainty on real data by 1) performing a cross-validation analysis of 521 genes in the
    ``essential'' Perturb-seq screen~\citep{Replogle2022} and 2) performing an out of distribution replication analysis
    between the ``essential'' and independent ``genome-wide'' Perturb-seq screen,
    showing that posterior inclusion probabilities (PIPs) provide robust and practical measures of edge significance across 
    folds and across datasets.
\end{itemize}

\section{Background}
\subsection{Causal discovery}
Causal discovery is the process of learning a causal graphical model (CGM), specifically a structural equation model (SEM),
on a set of variables from data~\citep{Pearl2009}. Formally, the model is represented by a DAG $\mathcal{G} = (V,E)$ where each
node $i\in V$ corresponds to one of $D$ variables $\{y_i\}_{i=1}^D$ and an edge $(i,j)\in E$ represents a direct effect of $y_i$ on $y_j$.
This is augmented by a set of functions detailing the relationship between each variable and its causal parents,
$y_i = f_i(y_{pa_i^\mathcal{G}}, \epsilon_i)$, where $\epsilon_i$ is an exogenous noise variable. Typically the variables
$\{\epsilon_i\}_{i=1}^D$ are assumed random, unmeasured and independent, and we can equivalently specify the model
in terms of probability distributions $p_i(y_i|y_{pa_i^\mathcal{G}})$. Assuming causal sufficiency, \textit{i.e.} that all other
relevant variables are measured, this gives the causal model.

The data, $\mathcal{D}$, can be either purely observational, consisting of $N$ measurements of the variables, $Y=[y_{ni}]_{n=1,i=1}^{N,D}$,
or interventional. With interventional data, the measured variables are augmented by a binary indicator matrix
$X=[x_{nm}]_{n=1,m=1}^{N,M}$, where $x_{nm}=1$ if sample $n$ received intervention $m$. These interventions can
be either \textit{hard} (structural) or \textit{soft} (shift). With hard interventions, incoming edges to the intervened-on variables are
assumed to be broken, yielding a set of interventional distributions $p_{m_i}(x_i)$ for each $i\in\mathcal{I}_m$. With
soft interventions, the distribution is changed without breaking the incoming connections, $p_{m_i}(x_i|y_{pa_i^G})$.
The targets of the interventions $\{\mathcal{I}_m\}$ can be assumed to be known or unknown.

\subsection{The linear-autoregressive causal model}
While inferring the parents of each node $pa_i^\mathcal{G}$ and the full form of $p_i$ for every variable $y_i$ is an
admirable goal, a more tractable goal in practice is to summarize the direct effect of one node on another
in terms of an average causal effect. This motivates the approximation 
$f_j(y_{pa_j^\mathcal{G}}, \epsilon_i) := \sum_i y_iG_{ij} + \epsilon_i$, where $G_{ij}$ is the direct effect size of $i$ on $j$.
This gives the well-studied linear-autoregressive causal model that we consider here
~\citep{geiger_parameter_2002,hyttinen_learning_2012,zheng2018dags,brown2023large,weinstock_gene_2024},
\begin{equation}
    Y = YG + \epsilon,
\label{eq:lar}
\end{equation}
with $G=[G_{ij}]_{i=1,j=1}^{D,D}$ the weighted adjacency matrix encoding the graph. 

\subsection{Bayesian causal discovery}
In Bayesian causal discovery, the goal is to estimate the entire posterior distribution of CGMs given the data,
$p(\mathcal{G}, \Theta | \mathcal{D})$, where $\Theta$ parameterizes the conditional distributions $p_i$~\citep{friedman_being_2003}.
This requires prior distributions over DAGs $p(\mathcal{G})$ and parameters $p(\Theta|\mathcal{G})$, and a likelihood $p(\mathcal{D}|\mathcal{G},\Theta)$, so that Bayes rule can be applied to yield the posterior. Under the linear auto-regressive model the prior and likelihood can be expressed in terms of the weighted adjacency matrix,
\begin{equation}
    p(G|\mathcal{D})\propto p(\mathcal{D}|G)p(G).
\end{equation}
The posterior encapsulates epistemic uncertainty regarding the structure of the graph given the data.
Due to the sheer number of possible graph structures in the posterior, we focus here on summarizing
the posterior through the posterior mean graph, which we define as the average weighted adjacency matrix
$\mathbb{E}[G|\mathcal{D}]$, and the posterior inclusion probability of each edge,
\begin{equation}
\mathrm{PIP}_{ij}
  \;=\;
  \Pr\!\bigl(|G_{ij}| > \varepsilon \,\big|\, \mathcal{D}\bigr),
  \label{eq:pip}
\end{equation}
where \(\varepsilon\) is a small edge–existence threshold.

% It enables calculation
% of the posterior mean graph, the posterior inclusion probability of individual edges, and the local false sign
% rate~\citep{stephens_false_2017}.
% of the posterior mean adjacency matrix and the posterior inclusion probability of individual edges.

\subsection{Instrumental variables and soft interventions}
When interventions are assumed to be hard, the interventional distribution is typically specified by removing
the dependency of the intervened node on its parents, $p_{m_i} = p_{\epsilon_i}$, or by fixing it to a specific value
such as 0, $\mathrm{do}(y_i=0)$. If the intervention instead results in a weak perturbation to its target,
the interventional distribution is more difficult to specify, but is commonly modeled as a mean shift~\citep{sani_identification_2020}. An alternative approach is to treat the interventions
as instrumental variables (IVs). $x$ is a valid IV for $y_i$ on $y_j$ if $x$ is not
independent of $y_i$ and if $x$ is independent of all exogenous factors affecting $y_j$ given $\mathrm{do}(
y_i)$~\citep{pearl_causality_2000}.
Thus in our model an intervention $\mathcal{I}_m$ targeting variable $y_m$ is trivially an IV for $y_m$
on every other variable $y_{i\neq m}$.

While there are many techniques for instrument variables 
analysis~\citep{levis_nonparametric_2024,imbens_instrumental_2014,baiocchi2014instrumental}, 
the simplest to consider in
the linear case given here is two-stage least squares (2SLS). In 2SLS, the average casual effect $\mathbb{E}[y_j|do(y_i)]=R_{ij}y_i$
is estimated by regressing $y_i$ on the instrument(s) $x$ to get $\hat{y}_i$, then regressing $y_j$ on $\hat{y}_i$,
\begin{equation}
    \hat{R}_{ij} = (Y_i^\top P_x Y_i)^{-1}P_x Y_i^\top Y_j
\end{equation}
with $P_x=x(x^\top x)^{-1}x^\top$ the projection matrix onto the column space of $x$. Note
that this measures the \textit{total} causal effect of $y_i$ on $y_j$ under the model $G$, not the direct
effect. Letting $\mathcal{P}_{i\rightarrow j}^G$ be the set of paths connecting $y_i$ and $y_j$
in $G$, we can write $R_{ij}$ as the sum of the product of the edge values in each path, $R_{ij} = \sum_{p \in \mathcal{P}_{i\rightarrow j}^G} \prod_{(k,l)\in p}G_{kl}$~\citep{hyttinen_learning_2012}.
\\

\noindent\textbf{Hard interventions.}
While our focus is on soft interventions, we note that
the same pipeline can accommodate hard interventions.
For an intervention \(\mathcal{I}_m\) with target \(y_i\), form \(G^{(m)}\) by zeroing column $m$ of \(G\) and write $Y^{(m)} = Y^{(m)} G^{(m)} + c^{(m)} + \varepsilon^{(m)}$.
In this regime \(y_i\) is exogenous, so the total effect on \(y_j\) is identified from the intervened samples by ordinary least squares, $\hat R_{ij} \;=\; \bigl({Y_i^{(m)}}^\top Y_i^{(m)}\bigr)^{-1} {Y_i^{(m)}}^\top Y_j^{(m)}$.
% These \(\hat R\) entries can replace the IV based estimates; see \citet{hyttinen_learning_2012} for details.

% \begin{figure}[t]
%   \centering
%   % scale the TikZ graphic to (up to) the column width
%   %\resizebox{\linewidth}{!}{%
%   \resizebox{0.5\linewidth}{!}{%
%   \tikz{
%     \node[latent] (pi) {$\pi_k^{i,j}$} ;
%     \node[latent, below = of pi] (sigma) {$\sigma_k^2$} ;
%     \node[latent, right = of pi, yshift=0.0cm] (G) {$G_{i,j}$} ;
%     \node[obs, below = of G, xshift=-0.7cm] (S) {$\hat{S}$} ;
%     \node[det, right = of G, yshift=0.0cm] (R) {$R$} ;
%     \node[obs, below = of R] (R_hat) {$\hat{R}$} ;
%     \node[det, right = of S, xshift=-0.5cm, yshift=0.5cm] (U) {$U$} ;
%     \node[det, below = of U, yshift=0.5cm] (V) {$V$} ;
%     {
%       \tikzset{plate caption/.append style={below right=0pt and 0pt of #1.south west}}
%       \plate[inner xsep=0.05cm, inner ysep=0.05cm]{plate1}{(G) (pi)} {$\;\;\quad\qquad\qquad\qquad D,D$} ;
%     }
%     \plate[inner sep=0.05cm]{plate2}{(sigma) (pi)} {$K$} ;
%     \edge{G}{S} ;
%     \edge{pi}{G} ;
%     \edge{sigma}{G} ;
%     \edge{G}{R} ;
%     \edge{R}{R_hat} ;
%     \edge{U}{R_hat} ;
%     \edge{V}{R_hat} ;
%     \edge{S}{U} ;
%     \edge{S}{V} ;
%   }}% end resizebox + tikz
%   \caption{The IBCD model. We treat $\hat{R}$ and $\hat{S}$ as observed and model $\hat{R}$ as coming
%   from a matrix normal distribution $\mathcal{MN}(R,U,V)$ with covariance matrices $U$ and $V$ determined
%   from $\hat{S}$. We place a spike–and–slab prior on $G$ which determines the mean matrix $R$ via~\eqref{eq:r_g_inv}.}
%   \label{fig:plate}
% \end{figure}

\label{headings}

\section{Interventional Bayesian Causal Discovery}

\subsection{Data model}
We augment the standard linear-autoregressive model~\eqref{eq:lar} with soft interventions,
\begin{equation}
Y = YG + X\beta + \epsilon,
\label{eq:gen_model}
\end{equation}
where Y and X are the $N \times D$ and $N \times M$ data and intervention design matrices as before,
$\beta$ is the $M \times D$ matrix of unknown intervention on gene effects, and $\epsilon$ is
an $N \times D$ exogenous noise matrix.

We assume that the direct targets of the interventions are known but that the effects of the interventions
are unknown. That is, we know which entries of $\beta$ are non-zero but not their values. This is a
reasonable assumption in Perturb-seq data where the CRISPR guide target is known and off-target effects
are minimal~\citep{Replogle2022}. We also assume that every feature has at least one instrument so that every element
in the matrix $\hat{R}$ can be estimated. We assume that the exogenous variables $\epsilon_i$ are independent with bounded variance, but make no other assumptions about their distributions $p_{\epsilon_i}$. $R$ can also be written recursively,
\begin{equation}
\begin{split}
  R &= \left[ \sum_{k} R_{ik}G_{kj} \right]_{i=1,j=1}^{D,D} \\
    &= \sum_{d=0}^{\mathrm{max}\{|p|, p \in \mathcal{P}^G\}} G^d = (I - G)^{-1}
\end{split}
\label{eq:r_g_inv}
\end{equation}

This together with collecting terms of~\eqref{eq:gen_model} and multiplying by the inverse yields the model,
 \begin{equation}
     Y = (X\beta + \epsilon)(I-G)^{-1} = X\beta R + \epsilon R.
     \label{eq:data_model}
 \end{equation}

\subsection{Empirical Bayesian model}
\label{sec:ebcd}

\begin{figure}[t]
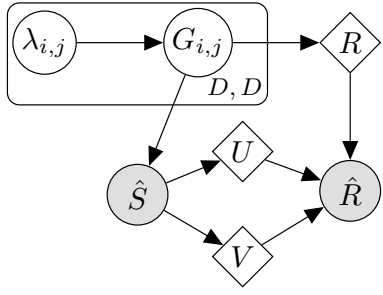

  \centering
  % scale the TikZ graphic to (up to) the column width
  %\resizebox{\linewidth}{!}{%
  \resizebox{0.3\linewidth}{!}{%
  \tikz{
    \node[latent] (lambda) {$\lambda_{i,j}$} ;
    %\node[latent, below = of pi] (sigma) {$\sigma_k^2$} ;
    \node[latent, right = of lambda, yshift=0.0cm] (G) {$G_{i,j}$} ;
    \node[obs, below = of G, xshift=-0.7cm] (S) {$\hat{S}$} ;
    \node[det, right = of G, yshift=0.0cm] (R) {$R$} ;
    \node[obs, below = of R] (R_hat) {$\hat{R}$} ;
    \node[det, right = of S, xshift=-0.5cm, yshift=0.5cm] (U) {$U$} ;
    \node[det, below = of U, yshift=0.5cm] (V) {$V$} ;
    {
      \tikzset{plate caption/.append style={below right=0pt and 0pt of #1.south west}}
      \plate[inner xsep=0.05cm, inner ysep=0.05cm]{plate1}{(G) (lambda)} {$\;\;\quad\qquad\qquad\qquad D,D$} ;
    }
    %\plate[inner sep=0.05cm]{plate2}{(sigma) (pi)} {$K$} ;
    \edge{G}{S} ;
    \edge{lambda}{G} ;
    %\edge{sigma}{G} ;
    \edge{G}{R} ;
    \edge{R}{R_hat} ;
    \edge{U}{R_hat} ;
    \edge{V}{R_hat} ;
    \edge{S}{U} ;
    \edge{S}{V} ;
  }}% end resizebox + tikz
  \caption{The IBCD model. We treat $\hat{R}$ and $\hat{S}$ as observed and model $\hat{R}$ as coming
  from a matrix normal distribution $\mathcal{MN}(R,U,V)$ with $U$ and $V$ determined from $\hat{S}$. We place a spike–and–slab horseshoe prior on $G$ which determines the mean matrix $R$ via~\eqref{eq:r_g_inv}.}
  \label{fig:plate}
\end{figure}

Our approach is inspired by a common strategy in genome-wide association studies,
where it is typically more computationally efficient to work with summary
statistics of the data, effect sizes and their standard errors, rather than the
full data matrix~\citep{Pasaniuc2017}. Thus, instead of modeling $Y$, we use $Y$ and $X$ to
estimate the total effects $\hat{R}_{ij}$ and their covariances 
$\hat{S}_{ijkl} = \mathrm{Cov}(\hat{R}_{ij}, \hat{R}_{kl})$. We then model
$\hat{R}$ with a matrix normal likelihood and place a horseshoe mixture prior on $G$.
See Figure~\ref{fig:plate} for an overview plate diagram.

\paragraph{Matrix normal likelihood.} If the instrumental variables
estimator has a normal distribution asymptotically, the likelihood 
$p\left(\hat{R}\mid G,\hat{S}\right)$ can be approximated by a matrix normal $\mathcal{MN}(R,U,V)=\mathcal{MN}((I-G)^{-1},U,V)$.
Recall that $X \sim \mathcal{MN}_{n \times p}(M, U, V)$,
if $\mathrm{vec}(X) \sim \mathcal{N}_{n\cdot p}(\mathrm{vec}(M), V \otimes U)$, where 
$\otimes$ denotes the Kronecker product. Here, $U \in \mathbb{R}^{n \times n}$ captures
row-wise correlations, $V \in \mathbb{R}^{p \times p}$ captures column-wise correlations,
and $S = V \otimes U$ defines the full $np \times np$ covariance matrix.
In our model, $U$ captures the covariances $\hat{S}_{i \cdot k \cdot}$
between interventions on variables induced by the design matrix $X$,
and $V$ captures the covariances $\hat{S}_{\cdot j \cdot l}$
between the response of variables to interventions. This
reduces the computational burden of estimating $D^4$ entries in the covariance matrix
$\hat{S}$ to estimating the $2D^2$ entries in the covariance matrices $U$ and $V$.

\paragraph{Estimating the matrix normal covariance matrices.}
% The matrix normal distribution provides a structured and computationally efficient alternative to the multivariate normal distribution for modeling matrix-valued random variables. Specifically, if $X \sim \mathcal{MN}{n \times p}(M, U, V)$, then $\mathrm{vec}(X) \sim \mathcal{N}{np}(\mathrm{vec}(M), V \otimes U)$, where $\otimes$ denotes the Kronecker product. Here, $U \in \mathbb{R}^{n \times n}$ captures row-wise correlations, $V \in \mathbb{R}^{p \times p}$ captures column-wise correlations, and $S = V \otimes U$ defines the full $np \times np$ covariance matrix. However, not all datasets naturally satisfy a Kronecker-structured covariance assumption, and directly estimating an unstructured $S$ is computationally infeasible in high dimensions. Additionally, the matrix normal likelihood offers a scalable approximation by decomposing $S$ into low-dimensional row and column covariances, reducing complexity from $O((np)^2)$ to $O(n^2 + p^2)$. In our case, we use this separable form to model uncertainty in the total effect matrix $\hat{R}$, where rows index perturbation targets and columns index gene responses.
\label{sec:res_cov}
For linear IV estimators, the entries of the variance-covariance matrix $\hat{S}$
are given by (Appendix~\ref{sec:sijkl_proof}),
\begin{equation}
    \hat{S}_{ijkl} = \frac{\mathrm{Cov}(\hat{y}_i,\hat{y}_k)\mathrm{Cov}(\hat{\epsilon}_{ij}, \hat{\epsilon}_{kl})}{n\mathrm{Var}(\hat{y}_i)\mathrm{Var}(\hat{y}_k)},
    \label{eq:s_ijkl}
\end{equation}
where $n$ is the total number of observations,
$\hat{y}_i$ is the projection of $y_i$ onto the space spanned by its instruments,
and $\hat{\epsilon}_{ij} = Y_j - \hat{R}_{ij}Y_i$ is an estimate of the residual.
Notice a natural separation into two terms, 
% \frac{\mathrm{Cov}(\hat{y}_i,\hat{y}_k)}{n\mathrm{Var}(\hat{y}_i)\mathrm{Var}(\hat{y}_j)}$,
the first representing covariation of the projections onto the instrument sets $X_i$ and $X_k$,
and % $\mathrm{Cov}(\hat{\epsilon}_{ij}, \hat{\epsilon}_{kl})$,
the second representing covariation
in the response of $y_j$ and $y_l$ to upstream interventions. While this response is 
dependent on the intervened variables, we average over interventions to approximate
the the matrix-normal column covariance. As such we can write,
\begin{align}
    \hat{U}_{ik} =& \frac{\mathrm{Cov}(\hat{y}_i,\hat{y}_k)}{n\mathrm{Var}(\hat{y}_i)\mathrm{Var}(\hat{y}_k)}, \\
    \hat{V}_{jl} =& \frac{1}{\sum_{ik} \mathbbm{1}\{U_{ik}\neq0\}} \sum_{i,k\mid U_{ik}\neq0} \mathrm{Cov}(\hat{\epsilon}_{ij}, \hat{\epsilon}_{kl}),
\end{align}
where $\mathbbm{1}\{U_{ik}\neq0\}$ indicates non-zero entries in $U$ which
are known through the intervention design matrix $X$. 

\begin{comment}
\paragraph{Fitting the global prior.}
We specify an empirical Bayes prior over $G$ informed by the structure
of the data, $p\left(G \mid \hat{S}\right)$. We assume each direct effect $G_{ij}$ follows a
unimodal distribution centered at zero. Following~\citep{stephens_false_2017}, we model this
using a mixture of a point mass at zero and a grid of zero-centered Gaussians, which offers
both flexibility and computational stability compared to traditional spike-and-slab or Laplace
priors. Our prior is given by,
\begin{equation}
\begin{split}
    p\!\left(G_{ij}\mid\{\pi_k\}, \{\sigma_k^2\}, \hat{s}_{ij}\right) 
    &\propto \pi_0 \, \mathcal{N}\!\left(0, \sigma_0^2 + \hat{s}_{ij}^2\right) \\
    &\quad + \sum_{k=1}^{K} \pi_k \, \mathcal{N}\!\left(0, \sigma_k^2 + \hat{s}_{ij}^2\right).
\end{split}
\end{equation}

where $\pi_0$ corresponds to a near-zero “spike” component % with fixed variance% $\sigma_0^2 = 10^{-3}$, 
and the slab weights $\{\pi_k\}$ are learned over a grid of variances $\{\sigma_k^2\}$.
The standard error of the estimate of $\hat{R}_{ij}$, given by $\hat{s}_{ij}=\hat{S}_{ijij}$,
accounts for edge-specific heteroskedasticity. We fit the parameters $\{\pi_k\}$
to the observed $\hat{R}$ using Expectation Maximization (EM), see Appendix~\ref{sec:appendix_a}
for details.
\end{comment}

\paragraph{Learning priors in sequence} 
We specify an empirical Bayes Spike and slab horseshoe prior over $G$ informed by the structure of the data, $p(G \mid \hat{S})$. We assume each direct effect $G_{ij}$ follows a unimodal distribution centered at zero. We first fit a global inclusion prior in the ER or SF form depending on structural assumptions to obtain marginal edge inclusion probabilities. We then localize these probabilities to edges and finally use them to parameterize the spike and slab horseshoe prior governing the edge weights.
\\
%We specify an empirical Bayesian Spike-and-slab horseshoe prior over $G$ informed by the structure of the data, $p(G \mid \hat{S})$.  We assume each direct effect $G_{ij}$ follows a unimodal distribution centered at zero. We consider both ER and SF prior distributions.
% We consider two approaches for constructing this prior: 
% (i) an Erdős–Rényi (ER) and 
% (ii) a Scale-Free (SF) priors.  

\noindent\textbf{Learning ER prior.}  
To capture ER structure, we estimate marginal edge inclusion probabilities with a mixture Gaussian to the observed effects $\hat{R}$~\citep{stephens_false_2017}. This flexible empirical Bayes prior captures both sparsity and heterogeneous effect sizes while remaining computationally stable compared to traditional spike and slab formulations. Details of the mixture specification and the Expectation Maximization (EM) estimation for the spike and slab weights are provided in Appendix~\ref{sec:appendix_a}.
\\
%To capture ER structure, we model each direct effect $G_{ij}$ using a mixture of a point mass at zero and a grid of zero-centered Gaussians~\citet{stephens_false_2017}. This flexible empirical Bayes prior captures both sparsity and heterogeneous effect sizes while remaining computationally stable compared to traditional spike-and-slab formulations. Details of the mixture specification and the Expectation Maximization (EM) estimation for the spike and slab mixture weights are provided in Appendix~\ref{sec:appendix_a}.

\begin{comment}
\begin{equation}
\begin{split}
    p\!\left(G_{ij}\mid\{\pi_k\}, \{\sigma_k^2\}, \hat{s}_{ij}\right) 
    &\propto \pi_0 \, \mathcal{N}\!\left(0, \sigma_0^2 + \hat{s}_{ij}^2\right) \\
    &\quad + \sum_{k=1}^{K} \pi_k \, \mathcal{N}\!\left(0, \sigma_k^2 + \hat{s}_{ij}^2\right),
\end{split}
\label{eq:slab_mixture}
\end{equation}
where $\pi_0$ corresponds to a near-zero “spike” component % with fixed variance% $\sigma_0^2 = 10^{-3}$, 
and the slab weights $\{\pi_k\}$ are learned over a grid of variances $\{\sigma_k^2\}$.
The standard error of the estimate of $\hat{R}_{ij}$, given by $\hat{s}_{ij}=\hat{S}_{ijij}$,
accounts for edge-specific heteroskedasticity. We fit the parameters $\{\pi_k\}$
to the observed $\hat{R}$ using Expectation Maximization (EM), see Appendix~\ref{sec:appendix_a}
for details.
\end{comment}

\noindent\textbf{Learning SF prior.}  
To capture SF structure, we derive edge inclusion probabilities from the squared total effect matrix $A = |\hat{R}|^2$, 
which amplifies strong signals and suppresses noise. 
We compute empirical out-weights $\theta_i = \sum_{j \ne i} A_{ij}$ and in-weights $\phi_j = \sum_{i \ne j} A_{ij}$, 
and solve the constrained optimization  

\begin{equation}
  \min_{P_{ij} \in [0,1],\, i \ne j} 
  \sum_{i=1}^D \left(\sum_{j \ne i} P_{ij} - \theta_i\right)^2 +
  \sum_{j=1}^D \left(\sum_{i \ne j} P_{ij} - \phi_j\right)^2
\end{equation}

subject to $P_{ii} = 0$, to recover a probabilistic edge inclusion matrix $P \in [0,1]^{D \times D}$. 
We then transform the solution into spike weights $\pi_{0,ij} = 1 - P_{ij}$.
\\
\begin{comment} 
we replace the mixture Gaussian slab in Eq.~\eqref{eq:slab_mixture} with a continuous, non-centered horseshoe prior, 
while retaining the spike component. 
The horseshoe prior is defined hierarchically as
$\theta_{ij} \sim \mathcal{N}(0, \tau^2 \lambda_{ij}^2), \quad \lambda_{ij} \sim \text{HalfCauchy}(0,1)$, where $\tau$ is a global shrinkage parameter and $\lambda_{ij}$ provides local adaptivity. Marginalizing out $\lambda_{ij}$ yields $p(\theta_{ij}\mid \tau) \;=\; 
    \int \mathcal{N}(0, \tau^2 \lambda_{ij}^2)\, p(\lambda_{ij})\, d\lambda_{ij}$,
which shows that the horseshoe can be viewed as a continuous scale mixture of Gaussians with infinitely many components. 

To further improve inference, we apply a non-centered reparameterization~\citep{papaspiliopoulos2007general} to the $\text{slab}_{ij} = \tau \cdot \lambda_{ij} \cdot \epsilon_{ij}$ with $\epsilon_{ij} \sim \mathcal{N}(0,1)$, $\lambda_{ij} \sim \text{HalfCauchy}(1)$, and fixed $\tau = 0.1$,
which decouples latent variables from their scales, avoids Neal’s funnel~\citep{neal2003slice}, and yields a smoother posterior for faster, more stable convergence. We show that the non-centered horseshoe slab replacement yields much faster and more stable inference in practice than Gaussian mixture slab 
(Eq.~\eqref{eq:slab_mixture}) in Table~\ref{tab:runtime}.
\end{comment}

\noindent\textbf{Localizing the global prior.}
\label{sec:learning_prior}
Assuming faithfulness~\citep{vonk_disentangling_2023}, two variables with a causal effect between them cannot be statistically independent.
As such, we capture the intuition that more correlated variables are more likely to reflect
a causal effect \textit{a priori} by building an edge specific prior, 
$p(G_{ij}) \propto \mathrm{Cov}(y_i, y_j)$ in the observational (non-intervened) data. Together, this
hybrid empirical Bayes strategy leverages both global patterns and local signal to construct a structured, data-driven prior over $G$ that improves both precision and interpretability in causal network inference, see Appendix~\ref{sec:appendix_a} for details.
\\

\noindent\textbf{Spike-and-slab horseshoe prior} 
After learning edge specific spike and slab weights separately from the ER and SF priors, we place a prior on each off-diagonal edge $G_{ij}$ in the form of a non-centered spike-and-slab horseshoe:
\begin{equation}
G_{ij} \sim \pi_{0,ij}\,\mathcal{N}(0,\sigma_0^2)\;+\;\bigl(1-\pi_{0,ij}\bigr)(\tau\,\lambda_{ij}\,\epsilon_{ij}), 
\end{equation}

where $\sigma_0^2 = 10^{-3}$, $\epsilon_{ij} \sim \mathcal{N}(0,1)$, $\lambda_{ij} \sim \mathrm{HalfCauchy}(1)$, and $\tau = 0.1$. We use the non-centered reparameterization~\citep{papaspiliopoulos2007general} $\theta_{ij}=\tau\,\lambda_{ij}\,\epsilon_{ij}$ of the hierarchical horseshoe $\theta_{ij} \sim \mathcal{N}(0, \tau^2 \lambda_{ij}^2)$, where $\tau$ is a global shrinkage parameter and $\lambda_{ij}$ provides local adaptivity. This reparameterization decouples latent variables from their scales and mitigates Neal’s funnel~\citep{neal2003slice} for faster and more stable convergence.

\paragraph{Sampling from the posterior.}
Together, these components define the posterior,

\begin{equation}
p\left(G \mid \hat{R}, \hat{S}\right) \propto p\left(\hat{R} \mid G, \hat{S}\right)  p\left(G \mid \hat{S}\right).
\label{eq:posterior}
\end{equation}

% Replacing the mixture Gaussian slab with the non-centered horseshoe prior smooths the complex posterior landscape and avoids Neal’s funnel pathology,
Our non-centered horseshoe prior produces a geometry that is well-suited for gradient-based inference. We perform inference using the No-U-Turn Sampler (NUTS) ~\citep{hoffman2014no}, implemented in NumPyro~\citep{bingham2019pyro, phan2019composable}, which leverages gradient information to scale inference to high-dimensional models, see Appendix~\ref{sec:appendix_inf} for details. 

\section{Experiments}
\label{experiment}

We performed comprehensive evaluations of our method as compared to others on both simulated and real data. For simulation, we generated datasets, with a known intervention design matrix, from the model~\eqref{eq:data_model} using ER and
SF DAGs varying dimensions, $D \in \{50,150,250,500\}$. Each dataset included $N=100$ intervention samples per variable and $D \times 100$ control samples, yielding
$2D \times 100$ total observations. This choice of $N=100$ conservatively matches the average of 141 cells per gene in the~\cite{Replogle2022} Perturb-seq experiment. In addition, for $D=50$ we conducted experiments with varying intervention sample sizes, $N \in \{5,15,25,50,75,100\}$, under the same design. For real data we analyzed normalized single cell Perturb seq expression matrices from K562 spanning the genome wide and essential gene screens. We retained 521 genes well-captured in both studies for downstream analysis after regressing out S and G2M cell cycle effects, see Appendix~\ref{sec:appendix_b} for details.
We compare the posterior mean weighted adjacency matrix inferred by IBCD, obtained by averaging marginal edge weights across posterior samples, against various methods
including NOTEARS~\citep{zheng2018dags}, SDCD~\citep{nazaret2023stable},
BayesDAG~\citep{annadani2023bayesdag}, LiNGAM~\citep{shimizu2006linear},
GIES~\citep{hauser2012characterization}, GES~\citep{chickering2002optimal},
inspre~\citep{brown2023large}, and LLCB~\citep{weinstock_gene_2024}. To ensure a fair comparison, we applied the same thresholding rule to binarize adjacency matrices across all methods, and for purely observational methods (GES, NOTEARS, LiNGAM, BayesDAG) we doubled the number of control samples to match the total sample size of interventional datasets. Further experiment details are provided in Appendix~\ref{sec:appendix_b}.

\begin{figure}[t]
  \centering
  \includegraphics[width=0.6\linewidth]{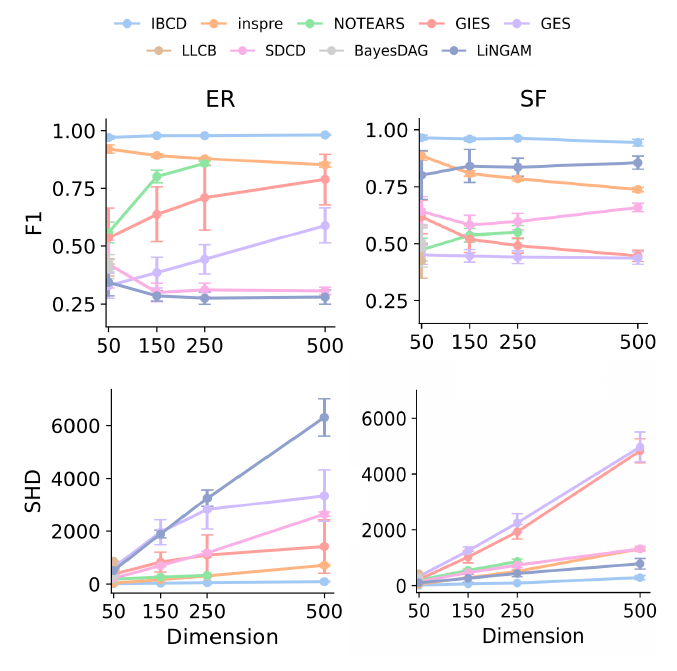}
  \caption{Comparison of F1 and SHD with increasing numbers of dimensions on ER and SF graphs, using 100 intervention samples per variable.}
  \label{fig:f1_shd_d}
\end{figure}

\subsection{Scaling with dimension on synthetic ER and SF graphs}

We evaluated each method by comparing the inferred graph to the ground truth using F1 score and structural Hamming
distance (SHD). We compared all methods on synthetic datasets with known ground truth graphs averaged over
ten replicates based on independently generated graphs and data. Figure~\ref{fig:f1_shd_d} summarizes F1 and SHD for $D\in\{50,150,250,500\}$ on ER and SF graphs.
%, using 100 intervention samples per $D$.
All optimization and score-based methods scaled to $D=500$ except NOTEARS which ran to $D=250$, while Bayesian methods LLCB and BayesDAG exceeded practical time or memory limits beyond $D=50$, see Table~\ref{tab:runtime} for details.
IBCD achieves the best overall performance, with the highest average F1 score and lowest average SHD in both ER and SF graphs, highlighting the benefit of learning a prior that adapts to the underlying graph topology. Detailed per-$D$ tables for F1, SHD, precision, and recall are provided in
Tables~\ref{tab:f1_d_details} and~\ref{tab:shd_d_details}, and
Figures~\ref{fig:f1_shd_d_precision} and~\ref{fig:f1_shd_d_recall} in
Appendix~\ref{sec:appendix_e}.

Regarding performance of other methods, NOTEARS and inspre perform better on ER graphs, consistent with theory that says $\ell_1$-regularized methods need more samples to recover hub-heavy SF topologies~\citep{liu2011reweighted}. By contrast, LiNGAM and SDCD are stronger on SF graphs. LiNGAM first estimates an ordering via independence tests, which we suspect may
be easier when a few hubs dominate, and SDCD’s spectral-radius penalty seems numerically more stable on SF structures. GIES, the interventional extension of GES, outperforms GES, highlighting the benefit of using interventional data, though on SF graphs the gain diminishes and can vanish as $D$ increases. We suspect that GES-family methods tend to do
better on ER than SF graphs because ER graphs have a more even degree distribution, so the score may change more smoothly with edge edits, yielding fewer near-ties and more reliable greedy steps. In contrast, SF hubs may create many colliders and dense neighborhoods, which yields many near equivalent choices and makes greedy search more prone to suboptimal moves. %Unless interventions hit those hubs, the extra orientation that GIES gains is diluted as the dimension grows.
More formal analysis would be required to validate this intuition.

\begin{figure}[t]
  \centering
  \includegraphics[width=0.6\linewidth]{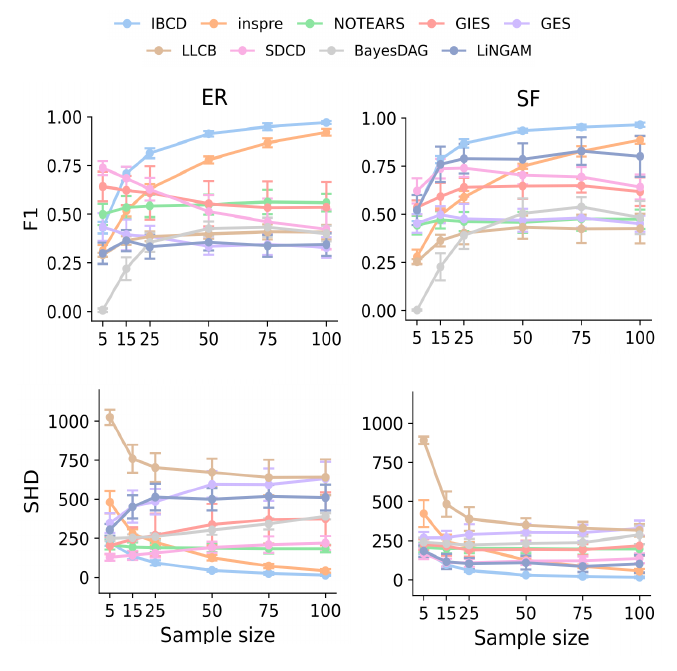}
  \caption{Comparison of F1 and SHD with increasing numbers of intervention sample sizes on ER and SF graphs with $D=50$.}
  \label{fig:f1_shd_n}
\end{figure}

\subsection{Scaling with sample size at $D=50$}
We again assess performance with F1 and SHD, now varying the number of intervention samples per variable $N$ at fixed dimension $D=50$ for both ER and SF graphs (Figure~\ref{fig:f1_shd_n}). As $N$ increases, most methods gain in F1 and drop in SHD with diminishing returns beyond about $N=50$. IBCD achieves the best overall performance on ER and SF graphs, achieving the highest mean F1 and lowest SHD in almost all settings, except the low-sample case (\(N=5\)). Detailed per-$N$ tables for F1, SHD, precision, and recall are provided in Tables~\ref{tab:f1_n_details} and~\ref{tab:shd_n_details}, and
Figures~\ref{fig:f1_shd_n_precision} and~\ref{fig:f1_shd_n_recall} in
Appendix~\ref{sec:appendix_e}.

\begin{table*}[t]
\centering
\label{tab:ablation_results}
\resizebox{\textwidth}{!}{%
\begin{tabular}{@{}lccccccccc@{}} % 1 l + 9 c = 10 columns total
\toprule
Variant & Empirical prior & MN & Edge specific & MVN & Global Prior & ER prior & SF prior & F1 & SHD \\
\midrule
ER baseline        & \checkmark & \checkmark & \checkmark & \(\times\) & \(\times\) & \checkmark & \(\times\) & 0.875 $\pm$ 0.023 & 58.9 $\pm$ 10.10 \\
Oracle Prior       & \(\times\) & \checkmark & \checkmark & \(\times\) & \(\times\) & \checkmark & \(\times\) & 0.876 $\pm$ 0.017 & 59.9 $\pm$ \,9.02 \\
MVN Likelihood     & \checkmark & \(\times\)  & \checkmark  & \checkmark  & \(\times\)  & \checkmark & \(\times\) & 0.431 $\pm$ 0.017 & 625.4 $\pm$ 18.75 \\
Global Prior       & \checkmark & \checkmark & \(\times\)   & \(\times\)   & \checkmark & \checkmark & \(\times\) & 0.845 $\pm$ 0.016 & 72.6 $\pm$ \,6.06 \\
ER with SF Prior   & \checkmark & \checkmark & \checkmark   & \(\times\)   & \(\times\)  & \(\times\) & \checkmark & 0.908 $\pm$ 0.017 & 44.5 $\pm$ \,8.34 \\
SF baseline        & \checkmark & \checkmark & \checkmark   & \(\times\)   & \(\times\)  & \(\times\) & \checkmark & 0.894 $\pm$ 0.020 & 47.6 $\pm$ \,8.44 \\
SF with ER Prior   & \checkmark & \checkmark & \checkmark   & \(\times\)   & \(\times\)  & \checkmark & \(\times\) & 0.809 $\pm$ 0.025 & 83.2 $\pm$ 11.42 \\
\bottomrule
\end{tabular}
}
\caption{Ablation study results comparing mean and standard deviation of F1 and SHD metrics across model variants averaged over 10 replicates of 50D ER graph. \checkmark\ indicates the presence of each component. MN indicates matrix normal, MVN indicates multivariate normal, ER indicates Erdős–Rényi, and SF indicates Scale-Free.}
\label{tab:ablation}
\end{table*}

\begin{table*}[t]
  \centering
  \scriptsize
  \setlength{\tabcolsep}{4pt}
  \resizebox{\textwidth}{!}{%
  \begin{tabular}{l*{11}{c}}
    \toprule
    Dim & Sample & IBCD  & inspre & GIES & GES & LiNGAM & SDCD & NOTEARS & BayesDAG & LLCB \\
    \midrule
    50D  & 5   & 2m 59s & 16s   & 1.3s & 0.6s & 1.4s  & 30s    & 37s    & 1m 52s & 10h 17m \\
    50D  & 15  & 3m 08s & 15s   & 0.8s & 0.9s & 2.6s  & 1m 17s & 53s    & 2m 49s & 10h 37m \\
    50D  & 25  & 3m 47s & 15.7s & 0.9s & 1.1s & 3.8s  & 2m 06s & 1m 14s & 3m 37s & 10h 41m \\
    50D  & 50  & 4m 11s & 14.9s & 2.1s & 1.4s & 6.9s  & 3m 49s & 2m 06s & 4m 58s & 11h 24m \\
    50D  & 75  & 4m 58s & 14.8s & 2.5s & 4.1s & 10.8s & 5m 17s & 3m 18s & 5m 59s & 10h 13m \\
    50D  & 100 & 6m 20s & 15.4s & 2.0s & 1.9s & 14.2s & 6m 46s & 3m 34s & 7m 07s & 10h 39m \\
    150D & 100 & 14m 30s & 4m 41s & 37s  & 46s  & 5m 03s & 26m 09s & 1h 10m & \textsc{NA} & \textsc{NA} \\
    250D & 100 & 24m 27s & 19m 41s & 2m 25s & 3m 28s & 22m 02s & 50m 05s & 2h 30m & \textsc{NA} & \textsc{NA} \\
    500D & 100 & 45m 38s & 2h 21m & 20m 15s & 41m 59s & 2h 55m & 1h 43m & \textsc{NA} & \textsc{NA} & \textsc{NA} \\
    \bottomrule
  \end{tabular}
  }
  \caption{Runtimes averaged across 10 replicates per setting, aggregated over ER and SF. All methods ran on a shared compute cluster with Intel Xeon E5-2660 v2 CPUs. IBCD and SDCD, the GPU-enabled methods, used NVIDIA H100 GPUs (80 GB VRAM each); IBCD leveraged three GPUs to parallelize MCMC sampling across chains. \textsc{NA} values correspond to failed runs, typically due to a 12 h timeout or termination for exceeding resource limits.}
\label{tab:runtime}
\end{table*}

\subsection{Ablation studies}
To assess the contribution of each modeling component, we conducted ablation studies 
comparing different configurations on $D$=50 synthetic ER and SF graphs in 
Table~\ref{tab:ablation} (see Appendix~\ref{sec:appendix_b}). The ER baseline learns an 
empirical prior from \(\hat{R}\) with a matrix normal likelihood and edge specific weights. 
An oracle prior that uses the true graph gives only a small, statistically insignificant 
gain, showing that the observational summary \(\hat{R}\) provides a strong empirical prior (see Appendix~\ref{sec:appendix_d}). 
To validate the likelihood we replaced the matrix normal with a multivariate normal that 
uses a truncated singular value decomposition of the covariance \(S\) keeping the top 
\(r=10\) components which reduces cost from \(\mathcal{O}(D^3)\) to \(\mathcal{O}(Dr^2)\) 
but reduced performance highlighting that learning structured row and column covariance 
matters. Using a single global prior also hurt which supports the value of edge specific 
priors obtained by localizing the global prior, highlighting the value of adaptive
sparsity. Surprisingly, using an SF prior on ER graphs improves performance at $D=50$
but degrades as $D$ grows to 150, 
250, and 500 with large drops in F1 and increases in SHD (Table~\ref{tab:ablation2}, Appendix~\ref{sec:appendix_e}). 
In contrast, the SF baseline outperforms SF with an ER prior, 
indicating that when the topology is unknown the SF baseline is a safer default.

\subsection{Posterior edge confidence diagnostics on 500 nodes graphs}
\begin{figure}[t]
  \centering
  \includegraphics[width=0.6\linewidth]{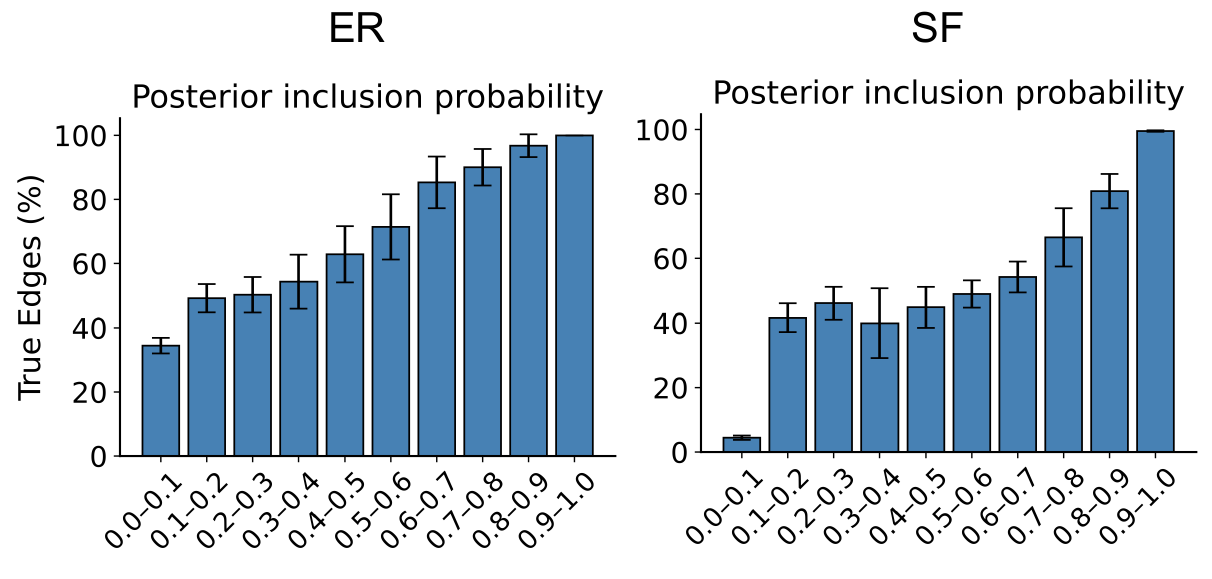}
  \caption{Calibration of posterior inclusion probability on 500D graphs under ER (left) and SF (right) graph in Figure~\ref{fig:f1_shd_d}. Bars show mean~$\pm$~s.d.\ over ten replicates. True edges concentrate in high PIP, confirming reliable uncertainty quantification.}
  \label{fig:pip_lfsr1}
\end{figure}

To verify that our $D=500$ inference is not only scalable but also trustworthy at the edge level, we analyzed posterior inclusion probability (PIP)~\eqref{eq:pip}, with $\epsilon = 0.05$.
The PIP quantifies posterior support for the presence of the edge. If PIPs are appropriately calibrated, approximately $80\%$ of edges with $\mathrm{PIP}=0.80$ represent
true direct causal effects. Figure~\ref{fig:pip_lfsr1} shows that our reported PIPs are conservative for both ER and SF graphs in high-dimensional settings. Thus, PIPs can be used to assess confidence in individual edges even in difficult inference regimes.

% \paragraph{Definitions.} For each ordered pair of nodes \((i,j)\) let \(G_{ij}\) be a posterior draw of the directed weight from \(i\) to~\(j\).
% Given the observed data set \(\mathcal{D}\) we define
% \[
% \mathrm{PIP}_{ij}
%   \;=\;
%   \Pr\!\bigl(|G_{ij}| > \varepsilon \,\big|\, \mathcal{D}\bigr)
% \]
% where \(\varepsilon\) is a small edge–existence threshold that we set to $\epsilon = 0.05$.
% Thus,
% \(\mathrm{PIP}_{ij}\) quantifies posterior support for the presence of the edge. If PIPs are appropriately calibrated, approximately $80\%$ of edges with $\mathrm{PIP}=0.80$ represent
% true direct causal effects. Figure~\ref{fig:pip_lfsr1} shows that our reported PIPs are conservative for both ER and SF graphs in high-dimensional settings. Thus, PIPs can be used to assess confidence in individual edges even in difficult inference regimes.

\subsection{Analysis of real Perturb‑seq data}

\begin{figure}[t]
  \centering
  \includegraphics[width=0.6\linewidth]{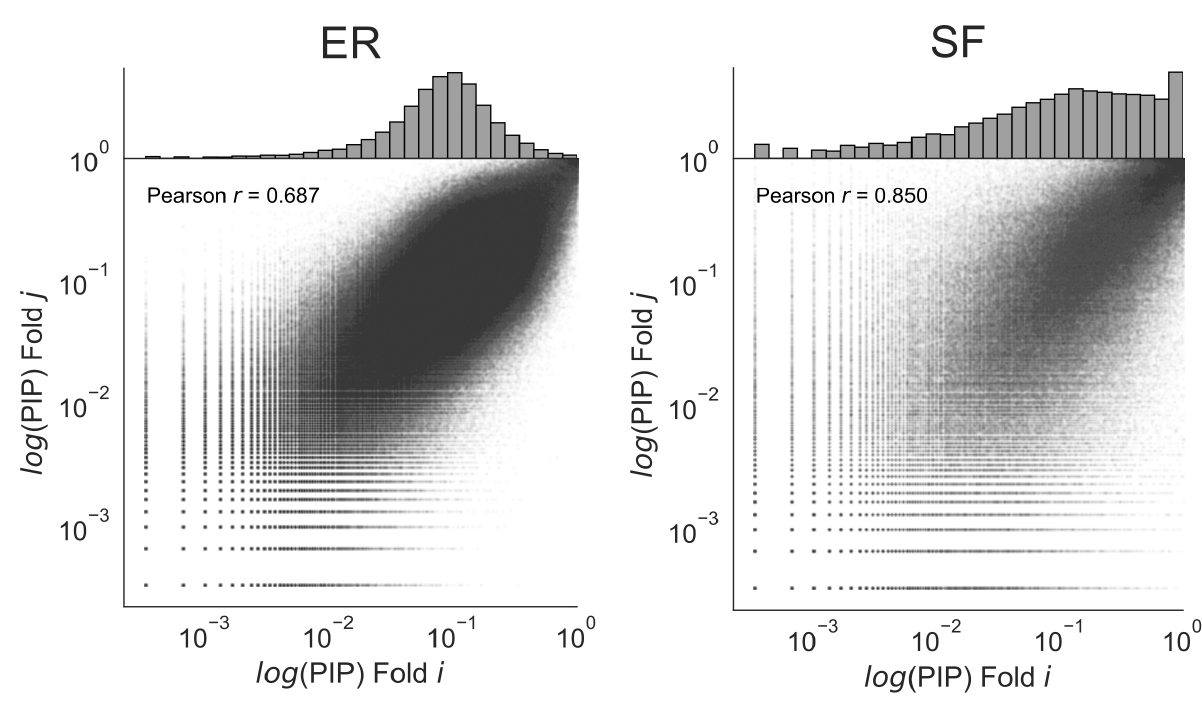}
  \caption{Agreement of log-transformed PIP across the ten cross-validation fold pairs on the K562 GWPS screen under ER and SF priors. Each point is one edge; the density of points forms a grey cloud. The positive correlation shows that high-PIP edges are consistently identified across folds.}
  \label{fig:pip_gwps}
\end{figure}

%We assessed IBCD on real data. 
Finally, we sought to assess the performance of IBCD on real data.
As there is a scarcity of large-scale data with ground truth causal graphs, we used five-fold stratified cross validation on 521 genes in the essential and genome-wide Perturb-seq screens~\citep{Replogle2022}. For each fold \(f\in\{1,\dots,5\}\), we fit the model on the training cells and computed \(\text{PIP}^{(f)}\) on the held-out cells. The scatter plot in Figures~\ref{fig:pip_gwps} and ~\ref{fig:pip_essential} overlays all ten unordered fold pairs \(\bigl(\text{PIP}^{(i)},\text{PIP}^{(j)}\bigr)\) (\(i<j\)) for the GWPS and essential screens, respectively. Each point is one putative edge, axes are on a log scale, and the top panel shows the marginal distribution of PIP values. Both figures show that the global Pearson correlation across the ten fold pairs is higher with the SF prior ($r = 0.850$) than with the ER prior ($r = 0.687$). This is consistent with hub-dominated, heterogeneous gene regulatory networks, so a SF prior better matches the data. Adapting the prior to the topology improves the reproducibility of edge-level PIPs across sub-samples and supports the reliability of our 521-gene inference.

\begin{figure}[H]
  \centering
  \includegraphics[width=0.6\linewidth]{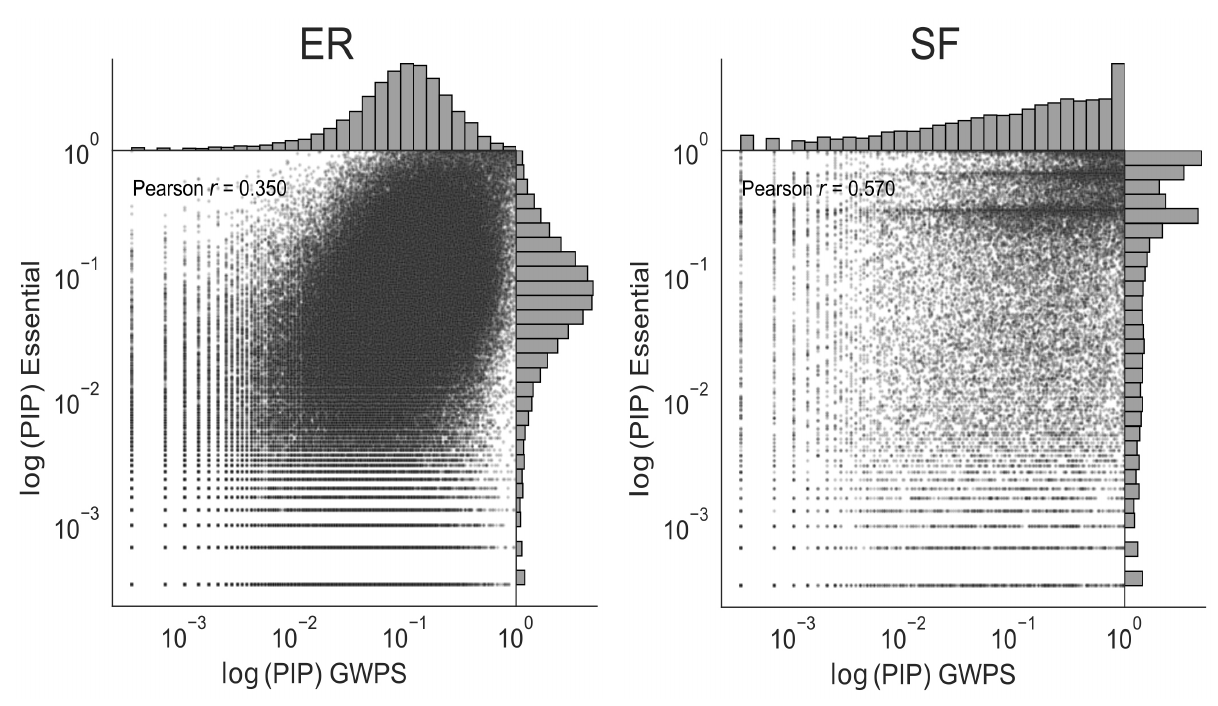}
  \caption{Agreement of log PIP values between GWPS and Essential K562 screens under ER and SF priors. Each point is one edge and darker regions indicate higher density. The SF prior shows stronger cross screen consistency.}
  \label{fig:gwps_essential}
\end{figure}

% \subsection{Out of sample validation experiment}
As the genome-wide and essential screens represent independent experiments with inevitable
differences in unmeasured exogenous factors, we conduced an out of distribution validation experiment comparing
the edge PIPs in both experiments. 
% Finally, we assessed uncertainty on real data using two K562 Perturb seq screens. Using inspre~\citep{brown2023large}, we first compared the two networks after restricting to the 521 shared genes and found an F1 near 0.2, likely reflecting differences in guide sets, efficiencies, depth, and cell state across screens, which underscores how brittle hard thresholded graphs can be.
We fit IBCD on each full dataset and obtained posterior inclusion probabilities for all edges. We then compared PIP from the genome wide screen with PIP from the essential screen and observed a strong positive correlation ($r=0.570$ for the SF prior, Figure~\ref{fig:gwps_essential}). Again the correlation is much higher with an SF prior than with an ER prior which is consistent with hub dominated gene regulation in these cells.
% These results underscore the value of uncertainty quantification since PIPs let us prioritize edges that replicate across screens and communicate edge level confidence instead of reporting a single hard graph.

\section{Discussion}
We have introduced Interventional Bayesian Causal Discovery (IBCD), an empirical Bayesian framework that scales
Bayesian causal inference to hundreds of nodes while providing calibrated edge level uncertainty. IBCD departs from most existing work in two key ways. First, we treat experimentally perturbed variables as instruments using a soft intervention framework and summarize the data by the matrix of total causal effects, allowing a matrix normal likelihood that reduces the computational burden of likelihood evaluation by orders of magnitude. Second, we learn a spike-and-slab horseshoe prior, with data-driven global sparsity and edge specific weights adapted to the local covariance structure in the observational samples. We also fit topology aware priors separately for ER and SF regimes, allowing the learned prior to match the graph family while using the same inference engine. Third, we show that PIP estimates are stable across cross validation folds and highly correlated across independent experiments, with stronger correlation under the SF prior. This indicates that IBCD yields reproducible edge probabilities on real interventional data at scale.

However, the present work has several limitations that point toward clear avenues for improvement. First, we make several assumptions. Perhaps the most limiting is the assumption of a linear model. While non-linearities are sure to exist, the linear model is well-studied and represents the classic tradeoff between expressivity and tractability. We also assume that every variable has at least one instrument. While these types of large-scale screens are becoming common in genomics, this limits applicability in scenarios where interventions are more difficult. Second, while we make no assumption on the distribution of error terms, we do assume that the IV estimator used to produce causal effect estimates is asymptotically normal. Relatedly, we focus on two-stage least-squares for IV estimates for simplicity, which assumes independence between the instrument and confounders that might not hold in practice. However, note that this framework allows any IV estimator with an asymptotic normal distribution to be used instead. Finally, our aim is to summarize the posterior via the mean graph and quantify edge level uncertainty through PIPs, rather than to characterize the full posterior.
This enables straightforward comparisons with non-Bayesian methods but leaves important questions regarding the modal distribution of the posterior open.
We hope to leverage multi-path approaches in future work to further characterize the posterior.

By providing scalable, uncertainty aware causal discovery, IBCD can help prioritize high confidence direct causal  effects. This can inform experimental follow-up while flagging those that remain ambiguous, thereby reducing costly false leads or guiding experimental design to clarify ambiguous connections. More broadly, we have demonstrated that the combination of instrumental variable summaries with empirical Bayes shrinkage is a powerful, efficient approach to causal discovery with high-dimensional intervention data.

\renewcommand{\bibname}{\Large References}
\begingroup
\normalsize
\bibliographystyle{apalike}
\bibliography{aistats,library_zot}
\endgroup

\clearpage
\appendix
\thispagestyle{empty}

% Supplementary material: To improve readability, you must use a single-column format for the supplementary material.

\section{Covariance structure of total causal effect estimates}
\label{sec:sijkl_proof}
Our goal is to estimate the covariance between the total causal effect estimates, given by
\begin{equation}
    \mathrm{Cov}(\hat{R}_{ij}, \hat{R}_{kl}) = \mathbb{E}[(\hat{R}_{ij} - \mathbb{E}[\hat{R}_{ij}])(\hat{R}_{kl} - \mathbb{E}[\hat{R}_{kl}])]
\end{equation}
We assume the estimator takes a two-stage linear form, such that the function linking the instruments to the exposure and the exposure
to the outcome are both linear:
\begin{align}
    \hat{y}_i =& P_i y_i, \\
    \hat{y}_j =& \hat{R}_{ij} y_i, \\
    \hat{R}_{ij} =& \hat{Q}_iy_j,
\end{align}
where we have written $P_i$ as the projection matrix onto the instruments for $i$, and $\hat{Q}_i$ as
the regression operator that determines the coefficients from $\hat{y}_i$. For example, in generalized least squares we would write
$P_i=X_i(X_i^\top W_i X_i) X_i^\top W_i$ and $\hat{Q}_i = (\hat{y}_i^\top W_y \hat{y}_i)^{-1}\hat{y}_i^\top W_y$ for weight
matrices $W_i$ and $W_y$. We assume the estimators are unbiased so that $\forall i,j,\; \mathbb{E}[\hat{R}_{ij}] = R_{ij}$.

Using~\ref{eq:data_model}, we can write $y_j$ as
\begin{equation}
    y_j = X\beta R_{:j} + \epsilon R_{:j},
\end{equation}
where $R_{:j}$ refers to the $j$'th column of $R$. Using this, we can write
\begin{equation}
    \hat{R}_{ij} = \hat{Q}_i(X\beta R_{:j} + \epsilon R_{:j}).
    \label{eq:rij}
\end{equation}
Similarly, writing $y_i$ as $y_iR_{ij} = X\beta R_{:i}R_{ij} + \epsilon R_{:i}R_{ij}$
and adding and subtracting $\hat{Q}_iy_i$ to~\ref{eq:rij} and collecting terms we have,
\begin{align}
    \hat{R}_{ij} &= \hat{Q}_i(X\beta R_{:j} + \epsilon R_{:j}) + \hat{Q}_iy_iR_{ij} - \hat{Q}_iy_iR_{ij} \\
        &= \hat{Q}_i(y_iR_{ij} + X\beta R_{:j} - X\beta R_{:i}R_{ij} + \epsilon R_{:j} - \epsilon R_{:i}R_{ij}) \\
        &= R_{ij} + \hat{Q}_i(X\beta R_{:j} - X\beta R_{:i}R_{ij} + \epsilon R_{:j} - \epsilon R_{:i}R_{ij}).
\end{align}
Because $\hat{Q}_i$ projects onto instruments for $y_i$, all effects $X\beta R_{:j}$ must travel through $y_i$,
so $\hat{Q}_iX\beta R_{:j} = \hat{Q}_iX\beta R_{:i}R_{ij}$. This leaves us with,
\begin{align}
    \hat{R}_{ij} - R_{ij} &= \hat{Q}_i(\epsilon R_{:j} - \epsilon R_{:i}R_{ij}) \\
        &= \hat{Q}_i( y_j - R_{ij}y_i ) \\
        &= \hat{Q}_i \epsilon_{ij}.
\end{align}

Using this and independence across samples we see that the covariance entries are given by
\begin{align}
    \mathrm{Cov}(\hat{R}_{ij}, \hat{R}_{kl}) &= \mathbb{E}[\hat{Q}_i \epsilon_{ij} \epsilon_{kl}^\top\hat{Q}_k^\top] \\
    &= \hat{Q}_i \hat{Q}_k^\top \mathrm{Cov}(\epsilon_{ij}, \epsilon_{kl}).
\end{align}
In the least squares case this leads to the plug in estimator~\ref{eq:s_ijkl},
\begin{equation*}
    \hat{S}_{ijkl} = \frac{\mathrm{Cov}(\hat{y}_i,\hat{y}_k)\mathrm{Cov}(\hat{\epsilon}_{ij}, \hat{\epsilon}_{kl})}{n\mathrm{Var}(\hat{y}_i)\mathrm{Var}(\hat{y}_k)}.
\end{equation*}

\section{Prior estimation}
\label{sec:appendix_a}

\subsection{ER global sparsity prior estimation}

As described in Section~\ref{sec:ebcd}, we place a flexible empirical Bayes prior on each direct effect $G_{ij}$.
Formally, the prior is
\begin{equation}
\begin{split}
    p\!\left(G_{ij}\mid\{\pi_k\}, \{\sigma_k^2\}, \hat{s}_{ij}\right) 
    &\propto \pi_0 \, \mathcal{N}\!\left(0, \sigma_0^2 + \hat{s}_{ij}^2\right) + \sum_{k=1}^{K} \pi_k \, \mathcal{N}\!\left(0, \sigma_k^2 + \hat{s}_{ij}^2\right),
\end{split}
\end{equation}
where $\pi_0$ corresponds to a near-zero "spike" component and the slab weights $\{\pi_k\}$ are learned over a grid of variances $\{\sigma_k^2\}$. The standard error of the estimate of $\hat{R}_{ij}$, given by $\hat{s}_{ij}=\hat{S}_{ijij}$, accounts for edge specific heteroskedasticity. We fit the parameters $\{\pi_k\}$ to the observed $\hat{R}$ using Expectation Maximization (EM).

The goal of the global sparsity prior is to estimate a flexible, data-driven mixture prior over all off-diagonal entries of the total effect matrix $\hat{R}$, capturing both sparsity and a range of plausible effect sizes. We require the following constraints:
\begin{equation}
    \pi_0 + \sum_{k=1}^{K} \pi_k = 1, \quad \pi_0 \geq 0, \quad \pi_k \geq 0.
\end{equation}
To estimate the mixture weights $\{\pi_0, \pi_1, \dots, \pi_K\}$, we maximize a penalized likelihood over the observed $\{w_i\}_{i=1}^N$. The log-likelihood of the data under this model is:
\begin{equation}
    \log L(\pi_0, \pi_1, \dots, \pi_K) = \sum_{i=1}^{N} \log \left[ \pi_0 f_0(w_i) + \sum_{k=1}^{K} \pi_k f_k(w_i) \right],
\end{equation}
%where $f_0(w) = \mathcal{N}(w; 0, \sigma_0^2)$ and $f_k(w) = \mathcal{N}(w; 0, \sigma_k^2)$. 

where $f_0(w_i) = \mathcal{N}(w_i; 0, \sigma_0^2 + \hat{s}_i^2)$ and $f_k(w_i) = \mathcal{N}(w_i; 0, \sigma_k^2 + \hat{s}_i^2)$ incorporate edge specific standard errors $\hat{s}_i$ from IV regression to reflect heteroskedasticity across edges. We add a Dirichlet-like penalty to the slab components to encourage sparsity:
\begin{equation}
    \text{Obj}(\pi_0, \pi_1, \dots, \pi_K) = \sum_{i=1}^{N} \log \left[ \pi_0 f_0(w_i) + \sum_{k=1}^{K} \pi_k f_k(w_i) \right] + (\alpha - 1) \sum_{k=1}^{K} \log \pi_k,
\end{equation}
where $\alpha = 2$ by default. A larger $\alpha$ concentrates more mass on the spike by shrinking the slab weights. We optimize this penalized objective using the EM algorithm. In the E-step, for each $w_i$, we compute the posterior responsibility for each mixture component:
\begin{align}
    z_{i,0} &= \frac{\pi_0 f_0(w_i)}{\pi_0 f_0(w_i) + \sum_{m=1}^{K} \pi_m f_m(w_i)}, \\
    z_{i,k} &= \frac{\pi_k f_k(w_i)}{\pi_0 f_0(w_i) + \sum_{m=1}^{K} \pi_m f_m(w_i)}.
\end{align}
We then compute expected counts:
\begin{equation}
    n_0 = \sum_{i=1}^{N} z_{i,0}, \quad n_k = \sum_{i=1}^{N} z_{i,k}, \quad k = 1, \dots, K.
\end{equation}
In the M-step, we update the mixture weights by maximizing the complete-data penalized log-likelihood:
\begin{align}
    \pi_0 &= \frac{n_0}{N + (\alpha - 1)K}, \\
    \pi_k &= \frac{n_k + (\alpha - 1)}{N + (\alpha - 1)K}, \quad k = 1, \dots, K.
\end{align}
These updates ensure that the weights remain normalized and non-negative. 

\subsection{Localizing ER and SF global sparsity prior estimation}

To localize the global prior, we calculate the covariance
$\Sigma=\mathrm{Cov}(Y_C)$ where $Y_C$ are the non-intervened control samples. We then
normalize this into a matrix of interaction strengths 
$\xi:= \frac{\Sigma \odot\Sigma}{||\Sigma\odot\Sigma||_F^2}$ such that each $\xi_{ij}\in[0,1]$.
We use this to construct edge specific weights $\pi_k^{ij}$ by solving the convex
quadratic program,
\begin{equation}
\min_{\pi_0, \pi_k} \sum_{i \ne j} \left[ \left( \pi_k^{i,j} - \xi_{ij} \right)^2 
  + \left( \pi_0^{i,j} - (1 - \xi_{ij}) \right)^2 \right],
\end{equation}
subject to the constraints,
\begin{align}
& \pi_0^{i,j} + \pi_k^{i,j} = 1 \quad \forall i \ne j, \\
& \pi_0^{i,i} = 1,\quad \pi_k^{i,i} = 0 \quad \forall i, \\
& \pi_0^{i,j} \ge 0,\quad \pi_k^{i,j} \ge 0 \quad \forall i,j,
\end{align}
with the sparsity constraint depending on the prior:  
\begin{align}
& \text{ER: } \sum_{i \ne j} \pi_0^{i,j} = \pi_0^{\text{global}} \cdot (D^2 - D), \\
& \text{SF: } \sum_{j > i} \pi_{0,ij} = \pi_{0,i} (D - i - 1) \quad \forall i,
\end{align}
where $\pi_0^{\text{global}}$ is the average sparsity level estimated from the EM step from ER, 
and $\pi_{0,i} \in [0,1]$ reflects the empirical sparsity of node $i$ from SF. We solve this convex optimization problem using the \texttt{CVXPY} package~\citep{diamond2016cvxpy}.

\section{Inference hyperparameters}
\label{sec:appendix_inf}

We used No-U-Turn Sampler (NUTS), with a target acceptance probability of 0.7, which balances step size and acceptance rate in Hamiltonian Monte Carlo, and a maximum tree depth of 10 to ensure thorough posterior exploration without excessive computation. The sampler is initialized at the median of three prior predictive draws, providing a stable starting point for sampling. We ran three chains with 300 warm-up steps and 1000 samples per chain.

\section{Experiment details}
\label{sec:appendix_b}

\subsection{Synthetic data generation and processing}

To evaluate performance across varied causal regimes, we used the simulator from~\citep{brown2023large}. We 
generated synthetic datasets with a known intervention design matrix from the model~\ref{eq:data_model} 
using both Erdős–Rényi (ER) and scale-free (SF) graph topologies while varying the graph size, $D \in 
\{50,150,250,500\}$. To keep sparsity comparable across sizes, we targeted an average node degree of 
approximately 5. For ER, we set the edge inclusion probability to $p \approx 5/D$, specifically  $p=\{0.10,
\,0.033,\,0.02,\,0.01\}$ for $D=\{50,150,250,500\}$. For SF, we empirically calibrated $p$ per $D$ to 
achieve an average degree of 5, selecting $p=\{0.066,\,0.108,\,0.124,\,0.139\}$ for $D=\{50,150,250,500\}$ 
via repeated simulation. Edge weights were drawn from a PERT distribution~\citep{clark1962pert} with minimum 
$v/2$, mode $v$, and maximum $2v$, where $v$ controls the typical magnitude of causal effects on a per-variance scale, and we set 
$v=0.25$. All data were normalized to mean 0, variance 1 in the control samples. To specify the strength of interventions, we defined the effect size $\beta$ as the number of 
standard deviations by which the mean of $Y_i$ shifts when the intervention $X_i = 1$ is applied. The 
intervention strength was fixed at $\beta=-2$. Each configuration included $D\times100$ interventional 
samples and $D\times100$ control samples ($2D\times100$ total). Additionally, for $D=50$ we created datasets 
with varying intervention sample sizes, $N\in\{5,15,25,50,75,100\}$, under the same design. For baselines 
that operate on observational data only (GES, NOTEARS, LiNGAM, BayesDAG), we generated purely observational 
datasets with $2D\times100$ control samples ($N=0$) so that the total sample size matches the interventional 
datasets. For ablations we set $\beta=-1.5$ and $v=0.2$ and used the same graph construction with ER graphs 
at $D\in\{50,150,250,500\}$ and SF graphs only at $D=50$ with $100$ intervention samples per variable and 
$D\times100$ control samples. All configurations were replicated across ten replicates to assess robustness 
of the methods.

\subsection{K562 Perturb-seq Data and processing}

%For the real data, we analyzed both the genome-wide (GWPS) and essential Perturb-seq screens in K562 cells~\citep{Replogle2022}, restricting to the 521 genes common to both panels. 

For real data, we analyzed both the genome-wide (GWPS) and essential Perturb-seq screens in K562 cells~\citep{Replogle2022}. We considered genes present in both panels and retained genes where the estimated per-variance effect-size of the intervention on the target gene was at least $-0.75$ standard deviations, and where there were at least 50 cells receiving a guide targeting that gene. This yielded 521 well-captured genes for downstream analysis.
Raw single-cell matrices were obtained from Figshare (\url{https://plus.figshare.com/ndownloader/files/35773075}) to regress out cell-cycle effects prior to downstream inference. This dataset contains CRISPR-inhibition guide RNA targeting each of 2,057 essential genes in the essential panel, and targeting approximately 9,866 expressed genes in the GWPS panels, with each cell receiving one guide targeting one gene. 

To enhance the perturbation signal and reduce cell cycle confounding, we regress out per-cell S and G2M module scores prior to inference. Let $Y\in\mathbb{R}^{C\times G}$ denote the log-normalized expression matrix after size-factor normalization, where $c$ indexes cells and $g$ indexes genes. Let $s_c$ be the S phase module score and $m_c$ the G2M module score, computed per cell from standard S and G2M gene sets with expression matched control genes. For each gene $g$, we fit
\[
Y_{cg} \;=\; \alpha_g \;+\; \beta^{(S)}_g\, s_c \;+\; \beta^{(\mathrm{G2M})}_g\, m_c \;+\; \varepsilon_{cg}.
\]
We define residuals using the fitted coefficients, $\hat\varepsilon_{cg} \;=\; Y_{cg} \;-\; \hat\alpha_g \;-\; \hat\beta^{(S)}_g\, s_c \;-\; \hat\beta^{(\mathrm{G2M})}_g\, m_c$.
Within each 10x GEM group $h$, we compute $\mu^{\mathrm{NTC}}_{gh}$ and $\sigma^{\mathrm{NTC}}_{gh}$ as the mean and standard deviation of $\hat\varepsilon_{cgh}$ over non-targeting control (NTC) cells in group $h$ and standardize residuals per gene using these NTC statistics within each group, $Z_{cgh} \;=\; \frac{\hat\varepsilon_{cgh} - \mu^{\mathrm{NTC}}_{gh}}{\sigma^{\mathrm{NTC}}_{gh}}$.
Here $Z_{cgh}$ denotes the standardized residual for cell $c$, gene $g$, and GEM group $h$. This yields a cell-cycle–adjusted and GEM-normalized matrix $Z$ for downstream analysis. We used two-stage least squares to estimate causal effects using guide presence as an instrumental variable. Note that
because each cell receives one guide, there is no covariance between $\hat{y}_i$ and $\hat{y}_k$ when $i\neq k$. Thus, the covariance~\ref{eq:s_ijkl} simplifies to
\[
\hat{S}_{ijkl} = 
\begin{cases}
\frac{1}{n \hat{\beta}_i^2}\,\mathrm{Cov}(\hat{\epsilon}_{ij},\hat{\epsilon}_{il}) & \text{if } i=k,\\[2pt]
0 & \text{otherwise.}
\end{cases}
\]
For each gene, we estimated the effect size $\hat{\beta}_i$ of the instrument on its target gene
using linear regression of gene expression on perturbation status. We then restricted analyses to the 521 genes shared by the GWPS and essential panels.

\subsection{Baselines}

To evaluate our method in a wider context, we compare our model (IBCD) to a diverse set of different methods  We describe our baselines briefly in the following.

\begin{itemize}
    \item \textbf{inspre~\citep{brown2023large}} inverse sparse regression (inspre) learns a sparse left inverse of the average causal effect matrix from per target interventions and reconstructs a weighted causal network. We use the implementation at \url{https://github.com/brielin/inspre} with default hyperparameters and set \texttt{DAG=TRUE} to enforce acyclicity.
    
    \item \textbf{LiNGAM~\citep{shimizu2006linear}} LiNGAM assumes linear relations, acyclicity, and independent non-Gaussian noise to recover a causal ordering and estimates edge weights by regression. We used the implementations from the \texttt{pcalg} R package with default parameter settings.
    
    \item \textbf{GES~\citep{chickering2002optimal}} Greedy Equivalence Search (GES) is a greedy score-based method for causal discovery using the BIC score. We used the implementations from the \texttt{pcalg} R package with default parameter settings.
    
    \item \textbf{GIES~\citep{hauser2012characterization}} Greedy Interventional Equivalence Search (GIES) is an interventional extension of GES~\citep{chickering2002optimal} that uses known intervention targets to score structures and orient edges. We used the implementations from the \texttt{pcalg} R with default settings. We include GIES to highlight the benefit of interventional data, and in our experiments it recovered graph structure better than the observational-only GES.
    
    \item \textbf{SDCD~\citep{nazaret2023stable}} Stable Differentiable Causal Discovery (SDCD) is a differentiable, score based method that enforces acyclicity with a spectral radius penalty and trains in two stages, edge preselection then penalized differentiable causal discovery, to improve stability and scale. We use the implementation at \url{https://github.com/azizilab/sdcd} with default parameters, setting \texttt{thresholding=True} to enable the built-in, data-adaptive thresholding driven by inferred effect sizes, following the jupyter notebook tutorial.

    \item \textbf{NOTEARS~\citep{zheng2018dags}} NOTEARS reformulates score-based DAG learning as a smooth equality-constrained optimization using a matrix-exponential acyclicity function, enabling global continuous search with standard solvers. We use the implementation at \url{https://github.com/xunzheng/notears} and set \texttt{w\_threshold}=0 to apply thresholding uniformly in post-processing. Our simulations use unit-normalized data such that edge weights represent per-variance effect sizes; therefore the default threshold 0.3 is well above the median simulated effect size and would degrade performance.
    
    \item \textbf{BayesDAG~\citep{annadani2023bayesdag}} BayesDAG combines hybrid strategy where MCMC learns permutations and mechanism parameters, while variational inference infers DAG edges given the learned permutations. We use the implementation at \url{https://github.com/microsoft/Project-BayesDAG} with default settings.
    
    \item \textbf{LLCB~\citep{weinstock_gene_2024}} Linear Latent Causal Bayes (LLCB) is a Bayesian extension of Linear Latent Causal model~\citep{hyttinen_learning_2012} that deconvolves total perturbation effects into direct gene–gene edges using sparsity and spectral-radius priors using a variational method, Pathfinder~\citep{zhang2022pathfinder}. We use the implementation at \url{https://github.com/weinstockj/LLCB} with default settings.

\end{itemize}

\section{Oracle priors from the true graph}
\label{sec:appendix_d}
In our ablation studies the oracle prior uses the true matrix \(G\) to directly set the spike probability and serves as an upper bound that isolates the impact of learning priors from observational summaries. We flatten the off diagonal entries of \(G\) into \(w=(w_1,\dots,w_N)\) with \(N=D(D-1)\). We compute
\[
\pi_0 = \frac{1}{N} \sum_{i=1}^{N} \mathbbm{1}\{w_i=0\}.
\]
The slab mass is \(1-\pi_0\). These weights parameterize the spike and slab horseshoe prior over \(G\) used in the ablation studies.

\section{Supplementary figures and tables}
\label{sec:appendix_e}

\begin{table}[h]
  \centering
  \resizebox{\textwidth}{!}{%
  \begin{tabular}{@{}llcccc@{}}
    \toprule
    Dim & metric & 50D & 150D & 250D & 500D \\
    \midrule
    \multirow{2}{*}{ER baseline}  
      & F1  & 0.875 $\pm$ 0.023 & 0.848 $\pm$ 0.019 & 0.858 $\pm$ 0.011 & 0.877 $\pm$ 0.015 \\
      & SHD & 58.9 $\pm$ 8.34  & 216.8 $\pm$ 25.36  & 335.9 $\pm$ 25.86  & 560.0 $\pm$ 56.47  \\
    \midrule
    \multirow{2}{*}{ER with SF prior} 
      & F1  & 0.908 $\pm$ 0.017 & 0.836 $\pm$ 0.022 & 0.789 $\pm$ 0.035 & 0.576 $\pm$ 0.131 \\
      & SHD & 44.5 $\pm$ 8.34  & 255.8 $\pm$ 33.19  & 598.0 $\pm$ 120.70 & 3160.8 $\pm$ 1502.28 \\
    \bottomrule
  \end{tabular}
  }
  \caption{Ablation study comparing the mean and standard deviation of F1 and SHD metrics between the ER baseline and ER with SF prior across different graph dimensions, averaged over ten replicates.}
\label{tab:ablation2}
\end{table}

\begin{table}[h]
  \centering
  \resizebox{\textwidth}{!}{%
  \begin{tabular}{@{}llccccccccc@{}}
    \toprule
    Dim & Graph & IBCD & inspre & GIES & GES & LiNGAM & SDCD & NOTEARS & BayesDAG & LLCB \\
    \midrule
    \multirow{2}{*}{50D}  & ER & 0.971 $\pm$ 0.008 & 0.921 $\pm$ 0.018 & 0.536 $\pm$ 0.129 & 0.329 $\pm$ 0.054 & 0.344 $\pm$ 0.058 & 0.422 $\pm$ 0.102 & 0.559 $\pm$ 0.046 & 0.400 $\pm$ 0.040 & 0.408 $\pm$ 0.037 \\
                          & SF & 0.965 $\pm$ 0.011 & 0.886 $\pm$ 0.020 & 0.617 $\pm$ 0.074 & 0.451 $\pm$ 0.042 & 0.801 $\pm$ 0.107 & 0.642 $\pm$ 0.065 & 0.474 $\pm$ 0.050 & 0.483 $\pm$ 0.086 & 0.426 $\pm$ 0.077 \\
    \midrule
    \multirow{2}{*}{150D} & ER & 0.978 $\pm$ 0.004 & 0.893 $\pm$ 0.008 & 0.638 $\pm$ 0.119 & 0.385 $\pm$ 0.066 & 0.285 $\pm$ 0.022 & 0.300 $\pm$ 0.040 & 0.801 $\pm$ 0.027 & \textsc{NA} & \textsc{NA} \\
                          & SF & 0.959 $\pm$ 0.009 & 0.808 $\pm$ 0.011 & 0.520 $\pm$ 0.059 & 0.447 $\pm$ 0.027 & 0.841 $\pm$ 0.072 & 0.582 $\pm$ 0.042 & 0.537 $\pm$ 0.029 & \textsc{NA} & \textsc{NA} \\
    \midrule
    \multirow{2}{*}{250D} & ER & 0.979 $\pm$ 0.003 & 0.879 $\pm$ 0.006 & 0.710 $\pm$ 0.140 & 0.443 $\pm$ 0.063 & 0.276 $\pm$ 0.026 & 0.311 $\pm$ 0.029 & 0.857 $\pm$ 0.010 & \textsc{NA} & \textsc{NA} \\
                          & SF & 0.963 $\pm$ 0.004 & 0.785 $\pm$ 0.011 & 0.492 $\pm$ 0.033 & 0.441 $\pm$ 0.028 & 0.836 $\pm$ 0.040 & 0.597 $\pm$ 0.036 & 0.551 $\pm$ 0.028 & \textsc{NA} & \textsc{NA} \\
    \midrule
    \multirow{2}{*}{500D} & ER & 0.982 $\pm$ 0.003 & 0.851 $\pm$ 0.010 & 0.789 $\pm$ 0.110 & 0.590 $\pm$ 0.076 & 0.280 $\pm$ 0.031 & 0.307 $\pm$ 0.016 & \textsc{NA} & \textsc{NA} & \textsc{NA} \\
                          & SF & 0.943 $\pm$ 0.015 & 0.738 $\pm$ 0.010 & 0.447 $\pm$ 0.025 & 0.437 $\pm$ 0.027 & 0.856 $\pm$ 0.029 & 0.659 $\pm$ 0.019 & \textsc{NA} & \textsc{NA} & \textsc{NA} \\
    \bottomrule
  \end{tabular}
  }
  \caption{Comparison of mean and standard deviation of F1 with increasing numbers of dimensions on ER and SF graphs in Figure~\ref{fig:f1_shd_d}. \textsc{NA} values correspond to failed runs, typically due to a 12 h timeout or termination for exceeding resource limits as noted in Table~\ref{tab:runtime}.}
\label{tab:f1_d_details}
\end{table}

\begin{table}[h]
  \centering
  \resizebox{\textwidth}{!}{%
  \begin{tabular}{@{}llccccccccc@{}}
    \toprule
    Dim & Graph & IBCD & inspre & GIES & GES & LiNGAM & SDCD & NOTEARS & BayesDAG & LLCB \\
    \midrule
    \multirow{2}{*}{50D}  & ER & 14.8 $\pm$ 4.26 & 41.3 $\pm$ 11.09 & 374.6 $\pm$ 168.58 & 632.1 $\pm$ 106.99 & 510.8 $\pm$ 82.95 & 219.9 $\pm$ 44.66 & 184.3 $\pm$ 24.53 & 391.7 $\pm$ 19.52 & 640.8 $\pm$ 114.74 \\
                          & SF & 16.6 $\pm$ 5.76 & 56.0 $\pm$ 12.87 & 218.0 $\pm$ 62.85  & 329.0 $\pm$ 49.67  & 101.9 $\pm$ 55.63 & 137.2 $\pm$ 30.98 & 196.6 $\pm$ 32.09 & 287.4 $\pm$ 58.90 & 316.4 $\pm$ 41.47 \\
    \midrule
    \multirow{2}{*}{150D} & ER & 32.0 $\pm$ 6.99 & 163.4 $\pm$ 15.15 & 829.7 $\pm$ 375.50 & 1958.8 $\pm$ 472.12 & 1889.4 $\pm$ 145.69 & 695.0 $\pm$ 40.04 & 266.3 $\pm$ 34.63 & \textsc{NA} & \textsc{NA} \\
                          & SF & 61.2 $\pm$ 13.65 & 279.7 $\pm$ 24.39 & 1024.9 $\pm$ 214.49 & 1247.2 $\pm$ 143.42 & 261.6 $\pm$ 143.87 & 462.8 $\pm$ 57.69 & 543.9 $\pm$ 53.11 & \textsc{NA} & \textsc{NA} \\
    \midrule
    \multirow{2}{*}{250D} & ER & 53.5 $\pm$ 8.58 & 308.9 $\pm$ 23.67 & 1094.3 $\pm$ 762.81 & 2819.5 $\pm$ 739.89 & 3254.1 $\pm$ 317.28 & 1178.5 $\pm$ 58.18 & 329.0 $\pm$ 20.62 & \textsc{NA} & \textsc{NA} \\
                          & SF & 91.7 $\pm$ 10.80 & 512.0 $\pm$ 43.91 & 1927.5 $\pm$ 261.34 & 2247.6 $\pm$ 328.42 & 435.5 $\pm$ 118.99 & 731.6 $\pm$ 85.21 & 862.1 $\pm$ 90.74 & \textsc{NA} & \textsc{NA} \\
    \midrule
    \multirow{2}{*}{500D} & ER & 91.5 $\pm$ 12.79 & 707.9 $\pm$ 28.25 & 1417.9 $\pm$ 1009.78 & 3348.2 $\pm$ 980.04 & 6306.1 $\pm$ 704.84 & 2638.6 $\pm$ 83.47 & \textsc{NA} & \textsc{NA} & \textsc{NA} \\
                          & SF & 284.8 $\pm$ 74.40 & 1316.5 $\pm$ 37.60 & 4831.8 $\pm$ 434.00 & 4972.1 $\pm$ 536.01 & 782.1 $\pm$ 190.47 & 1319.3 $\pm$ 88.99 & \textsc{NA} & \textsc{NA} & \textsc{NA} \\
    \bottomrule
  \end{tabular}
  }
  \caption{Comparison of mean and standard deviation of SHD with increasing numbers of dimensions on ER and SF graphs in Figure~\ref{fig:f1_shd_d}. \textsc{NA} values correspond to failed runs, typically due to a 12 h timeout or termination for exceeding resource limits as noted in Table~\ref{tab:runtime}.}
\label{tab:shd_d_details}
\end{table}

\begin{figure}[t]
  \centering
  \includegraphics[width=0.75\linewidth]{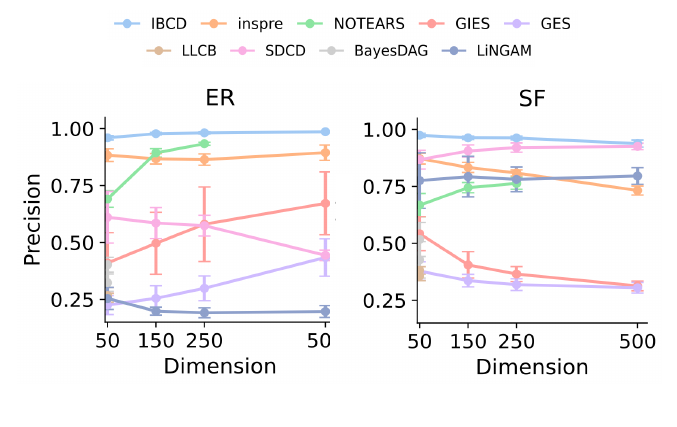}
  \caption{Comparison of precision with increasing numbers of dimensions on ER and SF graphs, using 100 intervention samples per variable.}
  \label{fig:f1_shd_d_precision}
\end{figure}

\begin{figure}[t]
  \centering
  \includegraphics[width=0.75\linewidth]{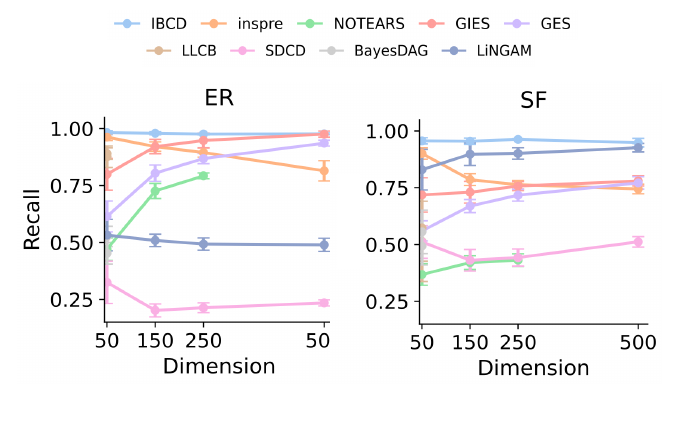}
  \caption{Comparison of recall with increasing numbers of dimensions on ER and SF graphs, using 100 intervention samples per variable.}
  \label{fig:f1_shd_d_recall}
\end{figure}

\begin{table}[h]
  \centering
  \resizebox{\textwidth}{!}{%
  \begin{tabular}{@{}llccccccccc@{}}
    \toprule
    Sample & Graph & IBCD & inspre & GIES & GES & LiNGAM & SDCD & NOTEARS & BayesDAG & LLCB \\
    \midrule
    \multirow{2}{*}{5}  & ER & 0.431 $\pm$ 0.030 & 0.311 $\pm$ 0.030 & 0.642 $\pm$ 0.077 & 0.434 $\pm$ 0.071 & 0.299 $\pm$ 0.054 & 0.737 $\pm$ 0.036 & 0.499 $\pm$ 0.055 & 0.007 $\pm$ 0.009 & 0.847 $\pm$ 0.023 \\
                        & SF & 0.518 $\pm$ 0.054 & 0.279 $\pm$ 0.037 & 0.536 $\pm$ 0.036 & 0.452 $\pm$ 0.049 & 0.522 $\pm$ 0.078 & 0.622 $\pm$ 0.065 & 0.443 $\pm$ 0.049 & 0.002 $\pm$ 0.004 & 0.633 $\pm$ 0.043 \\
    \midrule
    \multirow{2}{*}{15} & ER & 0.708 $\pm$ 0.035 & 0.517 $\pm$ 0.029 & 0.622 $\pm$ 0.075 & 0.394 $\pm$ 0.079 & 0.364 $\pm$ 0.054 & 0.684 $\pm$ 0.059 & 0.535 $\pm$ 0.058 & 0.220 $\pm$ 0.058 & 0.872 $\pm$ 0.026 \\
                        & SF & 0.784 $\pm$ 0.019 & 0.496 $\pm$ 0.035 & 0.592 $\pm$ 0.072 & 0.498 $\pm$ 0.044 & 0.760 $\pm$ 0.092 & 0.737 $\pm$ 0.051 & 0.471 $\pm$ 0.045 & 0.227 $\pm$ 0.070 & 0.586 $\pm$ 0.128 \\
    \midrule
    \multirow{2}{*}{25} & ER & 0.813 $\pm$ 0.026 & 0.630 $\pm$ 0.024 & 0.609 $\pm$ 0.137 & 0.387 $\pm$ 0.053 & 0.332 $\pm$ 0.061 & 0.628 $\pm$ 0.059 & 0.542 $\pm$ 0.057 & 0.355 $\pm$ 0.048 & 0.882 $\pm$ 0.025 \\
                        & SF & 0.868 $\pm$ 0.022 & 0.589 $\pm$ 0.032 & 0.640 $\pm$ 0.049 & 0.476 $\pm$ 0.078 & 0.789 $\pm$ 0.077 & 0.740 $\pm$ 0.046 & 0.462 $\pm$ 0.050 & 0.390 $\pm$ 0.070 & 0.564 $\pm$ 0.161 \\
    \midrule
    \multirow{2}{*}{50} & ER & 0.913 $\pm$ 0.014 & 0.779 $\pm$ 0.018 & 0.553 $\pm$ 0.116 & 0.334 $\pm$ 0.042 & 0.356 $\pm$ 0.041 & 0.513 $\pm$ 0.085 & 0.549 $\pm$ 0.052 & 0.426 $\pm$ 0.037 & 0.891 $\pm$ 0.025 \\
                        & SF & 0.934 $\pm$ 0.011 & 0.749 $\pm$ 0.013 & 0.647 $\pm$ 0.067 & 0.469 $\pm$ 0.060 & 0.786 $\pm$ 0.083 & 0.703 $\pm$ 0.048 & 0.458 $\pm$ 0.054 & 0.504 $\pm$ 0.078 & 0.573 $\pm$ 0.146 \\
    \midrule
    \multirow{2}{*}{75} & ER & 0.950 $\pm$ 0.018 & 0.867 $\pm$ 0.023 & 0.534 $\pm$ 0.135 & 0.344 $\pm$ 0.053 & 0.338 $\pm$ 0.057 & 0.461 $\pm$ 0.119 & 0.563 $\pm$ 0.063 & 0.433 $\pm$ 0.049 & 0.878 $\pm$ 0.043 \\
                        & SF & 0.953 $\pm$ 0.013 & 0.827 $\pm$ 0.027 & 0.649 $\pm$ 0.036 & 0.481 $\pm$ 0.050 & 0.828 $\pm$ 0.072 & 0.694 $\pm$ 0.051 & 0.476 $\pm$ 0.062 & 0.538 $\pm$ 0.051 & 0.532 $\pm$ 0.166 \\
    \midrule
    \multirow{2}{*}{100}& ER & 0.971 $\pm$ 0.008 & 0.921 $\pm$ 0.018 & 0.536 $\pm$ 0.129 & 0.329 $\pm$ 0.054 & 0.344 $\pm$ 0.058 & 0.422 $\pm$ 0.102 & 0.559 $\pm$ 0.046 & 0.400 $\pm$ 0.040 & 0.876 $\pm$ 0.047 \\
                        & SF & 0.965 $\pm$ 0.011 & 0.886 $\pm$ 0.020 & 0.617 $\pm$ 0.074 & 0.451 $\pm$ 0.042 & 0.801 $\pm$ 0.107 & 0.642 $\pm$ 0.065 & 0.474 $\pm$ 0.050 & 0.483 $\pm$ 0.086 & 0.514 $\pm$ 0.177 \\
    \bottomrule
  \end{tabular}
  }
  \caption{Comparison of mean and standard deviation of F1 with increasing numbers of intervention sample sizes on ER and SF graphs with $D = 50$ in Figure~\ref{fig:f1_shd_n}.}
\label{tab:f1_n_details}
\end{table}

\begin{table}[h]
  \centering
  \resizebox{\textwidth}{!}{%
  \begin{tabular}{@{}llccccccccc@{}}
    \toprule
    Sample & Graph & IBCD & inspre & GIES & GES & LiNGAM & SDCD & NOTEARS & BayesDAG & LLCB \\
    \midrule
    \multirow{2}{*}{5}  & ER & 225.3 $\pm$ 22.72 & 480.7 $\pm$ 71.30 & 203.5 $\pm$ 55.72 & 348.4 $\pm$ 62.84 & 304.6 $\pm$ 36.67 & 128.5 $\pm$ 21.66 & 208.7 $\pm$ 30.08 & 248.7 $\pm$ 16.57 & 1024.6 $\pm$ 48.99 \\
                        & SF & 186.9 $\pm$ 23.59 & 423.0 $\pm$ 85.45 & 225.3 $\pm$ 34.65 & 268.3 $\pm$ 36.97 & 183.5 $\pm$ 36.10 & 167.9 $\pm$ 34.80 & 206.8 $\pm$ 31.47 & 241.2 $\pm$ 27.53 & 891.3 $\pm$ 23.32 \\
    \midrule
    \multirow{2}{*}{15} & ER & 135.8 $\pm$ 19.32 & 280.3 $\pm$ 28.10 & 243.2 $\pm$ 77.52 & 452.8 $\pm$ 103.42 & 452.0 $\pm$ 73.57 & 141.5 $\pm$ 29.49 & 194.6 $\pm$ 29.35 & 253.9 $\pm$ 21.92 & 759.6 $\pm$ 89.35 \\
                        & SF & 96.3 $\pm$ 14.67  & 244.7 $\pm$ 43.44 & 214.3 $\pm$ 58.74 & 269.1 $\pm$ 43.67  & 116.1 $\pm$ 48.42 & 110.9 $\pm$ 25.36 & 198.6 $\pm$ 29.78 & 230.8 $\pm$ 29.37 & 483.9 $\pm$ 81.67 \\
    \midrule
    \multirow{2}{*}{25} & ER & 93.3 $\pm$ 16.42  & 225.2 $\pm$ 19.60 & 271.2 $\pm$ 136.74 & 484.6 $\pm$ 81.31  & 514.0 $\pm$ 85.27 & 155.3 $\pm$ 27.92 & 191.3 $\pm$ 30.22 & 261.1 $\pm$ 23.30 & 702.8 $\pm$ 92.57 \\
                        & SF & 59.3 $\pm$ 7.59   & 212.7 $\pm$ 29.58 & 190.4 $\pm$ 42.08  & 289.1 $\pm$ 67.28  & 103.3 $\pm$ 37.02 & 108.8 $\pm$ 25.94 & 202.0 $\pm$ 33.13 & 223.6 $\pm$ 37.25 & 389.2 $\pm$ 76.47 \\
    \midrule
    \multirow{2}{*}{50} & ER & 44.2 $\pm$ 8.28   & 126.0 $\pm$ 17.38 & 338.0 $\pm$ 132.32 & 595.6 $\pm$ 85.51  & 500.3 $\pm$ 71.75 & 192.6 $\pm$ 35.68 & 188.6 $\pm$ 27.82 & 303.5 $\pm$ 30.78 & 672.2 $\pm$ 87.79 \\
                        & SF & 30.5 $\pm$ 4.25   & 126.8 $\pm$ 16.64 & 193.5 $\pm$ 58.40  & 302.6 $\pm$ 56.32  & 109.5 $\pm$ 44.46 & 120.9 $\pm$ 25.94 & 200.2 $\pm$ 31.07 & 232.9 $\pm$ 48.99 & 350.0 $\pm$ 42.71 \\
    \midrule
    \multirow{2}{*}{75} & ER & 25.6 $\pm$ 8.68   & 72.5 $\pm$ 14.08  & 370.4 $\pm$ 154.47 & 594.0 $\pm$ 103.28 & 518.6 $\pm$ 73.58 & 209.4 $\pm$ 52.91 & 184.4 $\pm$ 32.34 & 342.6 $\pm$ 35.45 & 639.1 $\pm$ 113.90 \\
                        & SF & 22.2 $\pm$ 6.07   & 86.1 $\pm$ 12.05  & 191.9 $\pm$ 36.30  & 301.3 $\pm$ 51.68  & 85.4 $\pm$ 35.29  & 122.2 $\pm$ 27.58 & 196.8 $\pm$ 36.70 & 238.4 $\pm$ 37.65 & 330.6 $\pm$ 39.50 \\
    \midrule
    \multirow{2}{*}{100}& ER & 14.8 $\pm$ 4.26   & 41.3 $\pm$ 11.09  & 374.6 $\pm$ 168.58 & 632.1 $\pm$ 106.99 & 510.8 $\pm$ 82.95 & 219.9 $\pm$ 44.66 & 184.3 $\pm$ 24.53 & 391.7 $\pm$ 19.52 & 640.8 $\pm$ 114.74 \\
                        & SF & 16.6 $\pm$ 5.76   & 56.0 $\pm$ 12.87  & 218.0 $\pm$ 62.85  & 329.0 $\pm$ 49.67  & 101.9 $\pm$ 55.63 & 137.2 $\pm$ 30.98 & 196.6 $\pm$ 32.09 & 287.4 $\pm$ 58.90 & 316.4 $\pm$ 41.47 \\
    \bottomrule
  \end{tabular}
  }
  \caption{Comparison of mean and standard deviation of SHD with increasing numbers of intervention sample sizes on ER and SF graphs with $D = 50$ in Figure~\ref{fig:f1_shd_n}.}
\label{tab:shd_n_details}
\end{table}

\begin{figure}[t]
  \centering
  \includegraphics[width=0.75\linewidth]{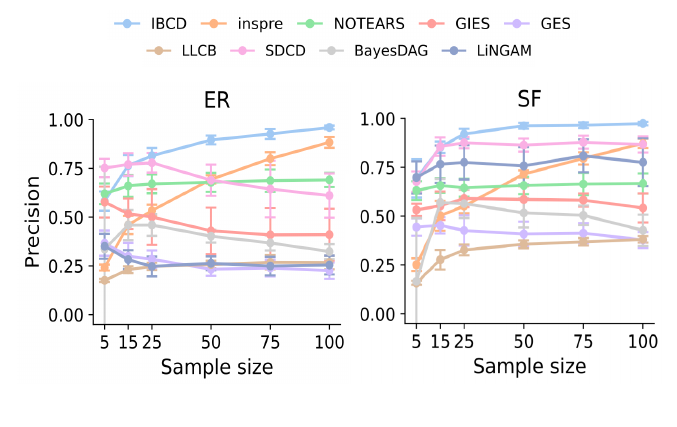}
  \caption{Comparison of precision with increasing numbers of intervention sample sizes on ER and SF graphs with $D=50$.}
  \label{fig:f1_shd_n_precision}
\end{figure}

\begin{figure}[t]
  \centering
  \includegraphics[width=0.74\linewidth]{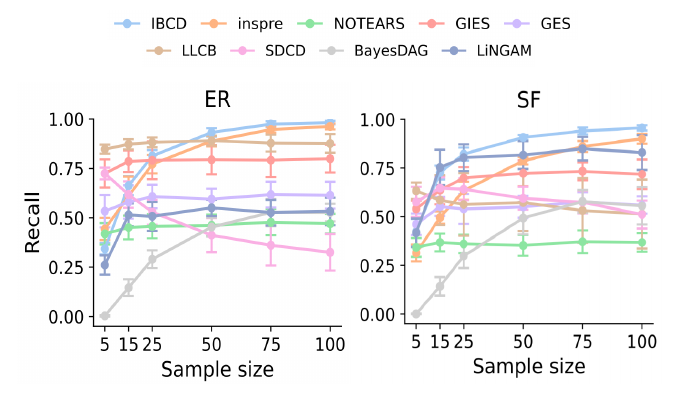}
  \caption{Comparison of recall with increasing numbers of intervention sample sizes on ER and SF graphs with $D=50$.}
  \label{fig:f1_shd_n_recall}
\end{figure}

\begin{figure}[t]
  \centering
  \includegraphics[width=0.75\linewidth]{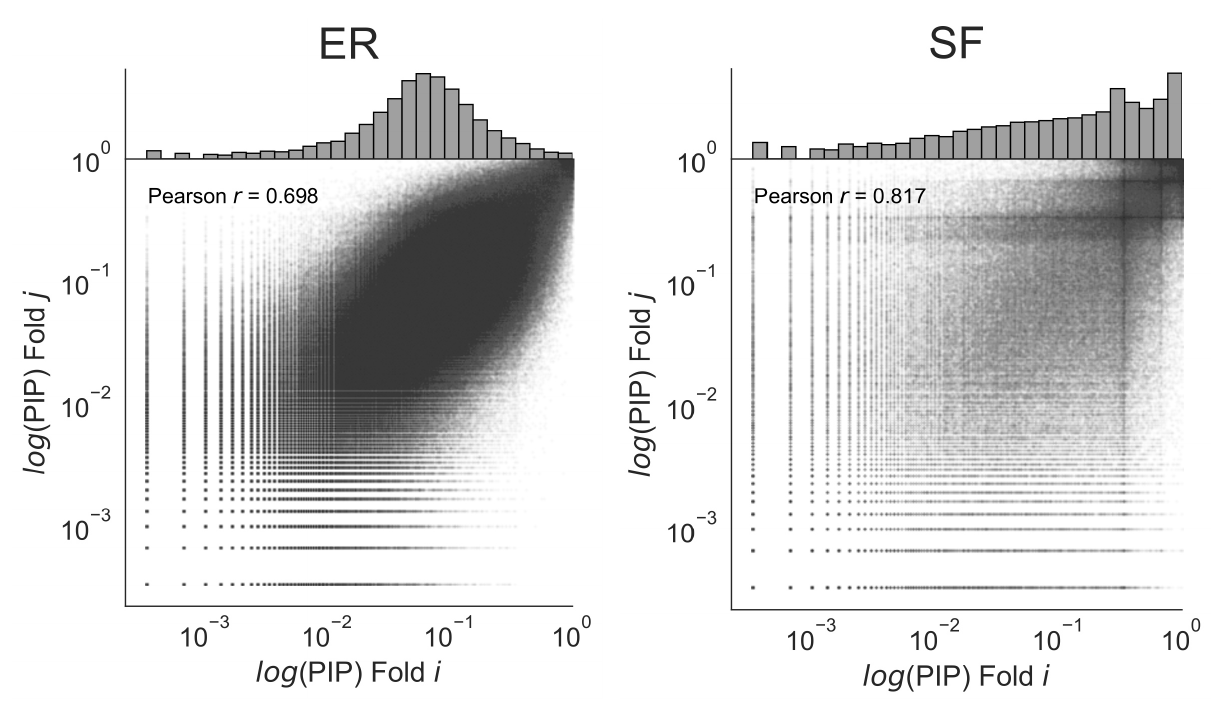}
  \caption{Agreement of log-transformed PIP across the ten cross-validation fold pairs on the K562 essential screen under ER and SF priors. Each point is one edge; the density of points forms a grey cloud. The positive correlation shows that high-PIP edges are consistently identified across folds.}
  \label{fig:pip_essential}
\end{figure}

\end{document}